\documentclass{article} 
\usepackage{preprint,times}

\usepackage{amsmath,amsfonts,bm}

\def\eqref#1{equation~\ref{#1}}

\def\1{\bm{1}}

\DeclareMathAlphabet{\mathsfit}{\encodingdefault}{\sfdefault}{m}{sl}
\SetMathAlphabet{\mathsfit}{bold}{\encodingdefault}{\sfdefault}{bx}{n}

\usepackage{hyperref}
\usepackage{url}

\usepackage{enumitem}
\usepackage{amsmath}
\usepackage{multirow}
\usepackage[most]{tcolorbox}
\tcbuselibrary{listings,breakable}
\usepackage{array}
\usepackage{paracol}
\usepackage{longtable}
\usepackage{ragged2e}
\usepackage{booktabs}

\newcolumntype{P}[1]{>{\RaggedRight\arraybackslash\footnotesize}p{#1}}

\newtcblisting{promptbox}[2][]{
  enhanced,
  breakable,
  listing only,
  colback=gray!5,
  colframe=gray!40,
  title={#2},
  listing options={
    basicstyle=\ttfamily\scriptsize,
    breaklines=true,
    breakatwhitespace=false,
    columns=fullflexible,
    keepspaces=true,
    showstringspaces=false
  },
  #1
}

\newcommand{\frameworkname}{DeferMem}  
\newcommand{\rlalgorithmname}{DistillPO}

\title{DeferMem: Query-Time Evidence Distillation via Reinforcement Learning for Long-Term Agent Memory}

\author{
Jianing Yin, Tan Tang, Yingcai Wu \\
State Key Lab of CAD\&CG, Zhejiang University \\
\texttt{yinjianing@zju.edu.cn}
}

\preprintfinalcopy
\begin{document}

\maketitle

\begin{abstract}
Large language model (LLM) agents still struggle to effectively use long-term memory, with answer-supporting evidence often scattered across long conversational histories and buried in substantial irrelevant content.
Existing memory systems commonly process memory into dedicated units before future queries are known and retrieve these preconstructed units at query time.
Because the contents and granularity of these units are determined before the current query is known, the retrieved units can remain coarse and noisy for that query, leaving downstream answerers to further denoise them and uncover query-specific evidence.
We present DeferMem, a long-term memory framework that decouples this problem into high-recall candidate retrieval and query-conditioned evidence distillation.
DeferMem uses a lightweight segment-link structure to organize raw history and retrieve broad candidates at query time.
A memory distiller then distills these high-recall but highly noisy candidates into a set of faithful, self-contained, and query-conditioned evidence.
To train this distiller, we introduce DistillPO, a reinforcement learning algorithm that formulates post-retrieval evidence distillation as a structured action comprising message selection and evidence rewriting.
It optimizes this action with a decomposed-and-gated reward pipeline and structure-aligned advantage assignment, gating reward components from validity to quality checks while exposing answerability-related feedback early and assigning each reward to its responsible output span.
On LoCoMo and LongMemEval-S, DeferMem surpasses strong baselines in QA accuracy and memory-system efficiency, achieving the highest QA accuracy and fastest runtime while consuming no commercial-API tokens for memory operations.
\end{abstract}

\section{Introduction}

Memory is essential for large language model (LLM) agents operating over long horizons, as it allows them to accumulate prior statements, track evolving states, retain user preferences, and recover evidence needed to answer later queries~\citep{huang2026survey_rethinking,hu2026survey_memory,wu2025survey_from,zhang2025survey}.
Although modern LLMs support increasingly large context windows~\citep{wang2024beyond_the_limits}, they struggle to effectively utilize long contexts due to attention dilution and the well-known ``lost in the middle'' problem~\citep{du2025survey_rethinking,liu2024lost}.
External memory systems therefore play a central role in enabling long-term interaction and reasoning~\citep{wu2025survey_from,du2025survey_rethinking}. 
However, leveraging long-term memory remains challenging, since the evidence needed for a query is often buried deep in history, fragmented across multiple sessions, and mixed with substantial irrelevant content~\citep{wu2025longmemeval,du2026memoryt1}.

Existing memory systems typically address this challenge through pre-query memory construction and query-time retrieval~\citep{zhong2024memorybank,pan2025secom,chen2025compress,kang2025memoryos,xu2026amem,fang2026lightmem}. 
Before future queries arrive, they often transform raw conversation histories into dedicated memory structures and maintain them through compression, forgetting, or updating mechanisms.
While such write-time memory construction improves storage efficiency and retrieval structure, it commits to a query-agnostic view of what matters and may discard or blur details that later become crucial for answering a specific query.
Even when raw history is retained, query-time memory retrieval commonly operates over preconstructed memory units.
Many systems rely on indirect relevance cues such as embedding similarity or keyword overlap~\citep{du2025survey_rethinking,jia2025survey_hippocampus,huang2026survey_rethinking}, while recent approaches move beyond one-shot similarity retrieval through candidate selection~\citep{tan2025prospect,yan2026memoryr1,du2026memoryt1,xu2026memgas} or iterative and agentic retrieval of memory units~\citep{zhang2025associa,yan2025gam,shu2026remem,hu2026evermemos,ji2026mragent}.
Although these advances make memory retrieval more query-adaptive, even retrieved units useful for answering the current query can remain coarse-grained and noisy and are not necessarily expressed in a query-conditioned form.
This leaves the downstream answerer to further denoise them and uncover query-specific evidence.
Therefore, distilling query-conditioned evidence from noisy retrieved memories remains a critical bottleneck.

\begin{figure}[t]
    \centering
    \includegraphics[width=\linewidth]{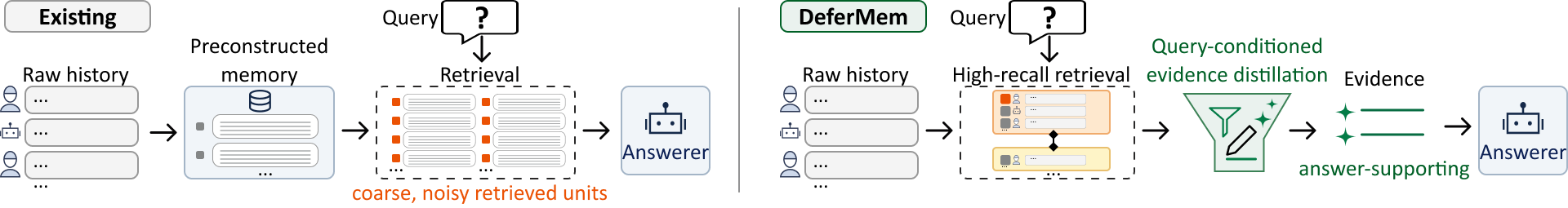}
    \caption{Comparison between existing memory systems (left) and DeferMem (right).}
    \label{fig:intro_example}
\end{figure}

To address this bottleneck, we propose a long-term memory framework, namely \frameworkname{}, which defers evidence distillation to query time.
Before future queries are known, \frameworkname{} uses a lightweight segment-link structure to preserve raw conversation history without replacing it with compressed memories.
When a query arrives, \frameworkname{} decouples memory utilization into two stages: high-recall candidate retrieval and query-conditioned evidence distillation.
First, it retrieves candidates by embedding similarity and expands them through segment links, yielding a high-recall but highly noisy candidate set.
Second, \frameworkname{} introduces a memory distiller that distills these broad candidates into a small set of faithful, self-contained, query-conditioned evidence for downstream answering.
To train this distiller, we introduce \rlalgorithmname{}, a reinforcement learning algorithm that formulates post-retrieval evidence distillation as a structured action consisting of useful-message selection and evidence rewriting.
Training such a distiller is challenging because the end-task reward alone is too coarse. 
To address this, \rlalgorithmname{} combines decomposed reward components with a leaky hierarchical gating strategy that follows the dependencies among rewards while ensuring sustained activation of answerability-related signals.
It further adopts structure-aligned advantage assignment to assign each reward signal to the output span responsible for it.
Together, these designs allow \frameworkname{} to retain raw history for high-recall retrieval while learning to distill noisy retrieved candidates into evidence ready for downstream answering.

Our contributions are summarized as follows:
\begin{itemize}[nosep,leftmargin=*]
    \item We introduce \rlalgorithmname{}, a reinforcement learning algorithm for post-retrieval evidence distillation. \rlalgorithmname{} trains a memory distiller to transform highly noisy candidates into a compact set of query-conditioned evidence by combining structured actions with decomposed rewards, leaky hierarchical reward gating, and structure-aligned advantage assignment.
    \item We propose \frameworkname{}, a long-term memory framework that preserves raw history in a lightweight segment-link structure and decouples query-time memory utilization into high-recall candidate retrieval and query-conditioned evidence distillation.
    \item We conduct extensive experiments on long-term memory benchmarks to evaluate \frameworkname{} from multiple perspectives. The results show that \frameworkname{} substantially improves question answering (QA) accuracy over strong baselines while achieving lower runtime and using no commercial-API tokens for memory operations.
\end{itemize}
\section{Related Work}

\textbf{Memory Systems for LLM Agents.}
Memory systems enable LLM agents to conduct long-term interactions~\citep{huang2026survey_rethinking,hu2026survey_memory,wu2025survey_from,zhang2025survey}. 
Prior work has explored dedicated storage structures, including graphs~\citep{jin2024graphcot,edge2024graphrag,guo2025lightrag,chhikara2025mem0,gutierrez2025hipporag2,rezazadeh2025memtree}, structured summaries or abstractions~\citep{zhong2024memorybank,li2025structrag,wang2025recuersum,chen2025compress,sun2026hmem}, linked notes~\mbox{\citep{xu2026amem}}, semantic segments~\mbox{\citep{pan2025secom,fang2026lightmem}}, multi-granularity memory associations~\citep{xu2026memgas}, and operating-system-style architectures~\citep{packer2024memgpt,kang2025memoryos}. 
Many systems further use compression~\citep{lee2024readagent,chen2025compress}, forgetting~\citep{zhong2024memorybank}, or updating~\citep{tan2025prospect,ong2025towards,li2025hello,xu2026amem} to maintain long-term memories.
While these pre-query designs make the storage more maintainable, they may discard details that later become crucial for a specific query.
At query time, many systems retrieve preconstructed memory units via embedding similarity, keyword matching, or metadata filtering~\citep{du2025survey_rethinking,jia2025survey_hippocampus,huang2026survey_rethinking}.
Recent systems augment one-shot similarity retrieval with candidate selection~\citep{tan2025prospect,yan2026memoryr1,du2026memoryt1,xu2026memgas} or iterative and agentic retrieval of memory units~\citep{zhang2025associa,yan2025gam,shu2026remem,hu2026evermemos,ji2026mragent}.
These methods improve the ability to surface memory units that are useful for the current query, but the returned units may still contain much more information than is needed to answer it, and that information may not be organized in a form directly tailored to the query, leaving downstream answerers to further uncover the answer-supporting information.

\textbf{Reinforcement Learning for LLMs.}
Reinforcement learning (RL) has become a central post-training paradigm for enhancing the native capabilities of LLMs~\citep{sang2025survey_beyondpipelines,zhang2026survey_landscape}.
Early reinforcement learning from human feedback methods commonly used PPO~\citep{schulman2017ppo} to train LLMs to generate outputs aligned with human preferences.
The development of algorithms such as GRPO~\citep{shao2024grpo} and DAPO~\citep{yu2026dapo} has enabled the training of LLMs for long-horizon tasks, particularly those characterized by long-term dependencies and sparse rewards.
These advances have enabled RL not only to increase the pass@1 success rate of LLMs on target tasks~\citep{yue2026does}, but also to elicit reasoning behaviors not readily observed in base LLMs~\citep{wen2026reinforcement}.
Recent work applies RL to a growing range of capabilities, including reasoning~\citep{guo2025deepseekr1}, tool use~\citep{feng2025retool}, web search~\citep{jin2025searchr1}, and memory~\citep{yan2026memoryr1,du2026memoryt1,tan2025prospect}. 
However, RL for effective long-term memory utilization remains relatively underexplored.
Prior memory-oriented RL methods mainly formulate this problem as selecting useful items from a given set of memory items, whereas \rlalgorithmname{} learns a structured policy that transforms noisy candidates into a compact set of query-conditioned evidence through message selection and evidence rewriting.
\section{DeferMem}

\subsection{Problem Formulation}

We study long-term memory QA for LLM agents. In this setting, an agent has access to a raw interaction history $H$ accumulated over a long horizon, either from its own interactions with a user or from conversations among other participants. When a new query $q$ arrives, the goal is to produce an answer $\hat{a}$ according to information contained in $H$.
We represent the raw interaction history as a sequence of sessions $H=\{S_1,S_2,\dots,S_n\}$, where each session $S_i = \{m_{i,1}, m_{i,2}, \dots, m_{i,|S_i|}\}$ consists of an ordered sequence of messages. 
Each message is represented as $m_{i,j}=(r_{i,j},c_{i,j},t_{i,j})$, where $r_{i,j}$, $c_{i,j}$, and $t_{i,j}$ denote the speaker identity, message content, and timestamp, respectively.
The content $c_{i,j}$ may include text and, when an image is attached, a textual description of the image.

Given a query $q$ and the history $H$, the task of a long-term memory system is to help the LLM agent generate an answer $\hat{a}$. 
Achieving this goal requires the system to locate and expose answer-supporting evidence embedded in $H$.
A central challenge is that the evidence required to answer $q$ is often sparse, fragmented, and buried in substantial irrelevant content. 
Let $E^*$ denote the latent set of answer-supporting evidence for $q$. 
In practice, $E^*$ may span multiple sessions, consist of only a few message fragments, and require information completion across multiple messages. 
As a result, directly answering from $H$ is inefficient, while recall-oriented retrieval alone typically returns a much larger candidate set $\hat{C} \subseteq H$ that contains $E^*$ but also contains substantial noise.

We therefore decompose query-time memory utilization into two stages:
\begin{equation}
\label{eq:two_stage_memory}
\hat{C} = \mathcal{R}(q, H), \qquad \hat{E} = \mathcal{D}(q, \hat{C}),
\end{equation}
where $\mathcal{R}$ is a high-recall retriever and $\mathcal{D}$ is a query-conditioned evidence distiller. The retrieved candidate set $\hat{C}$ is expected to contain the answer-supporting information with high recall, while the distilled evidence $\hat{E}$ should be a compact set of faithful, self-contained, and query-conditioned evidence that is directly useful for answering $q$. The downstream LLM agent can then produce the final answer $\hat{a}$ based on $q$ and $\hat{E}$. Under this formulation, the key problem addressed in this work is not only whether the system can retrieve relevant candidates, but also whether it can distill the right evidence for the current query from high-recall yet highly noisy candidates.

\begin{figure*}[!t]
  \centering 
  \includegraphics[width=\linewidth]{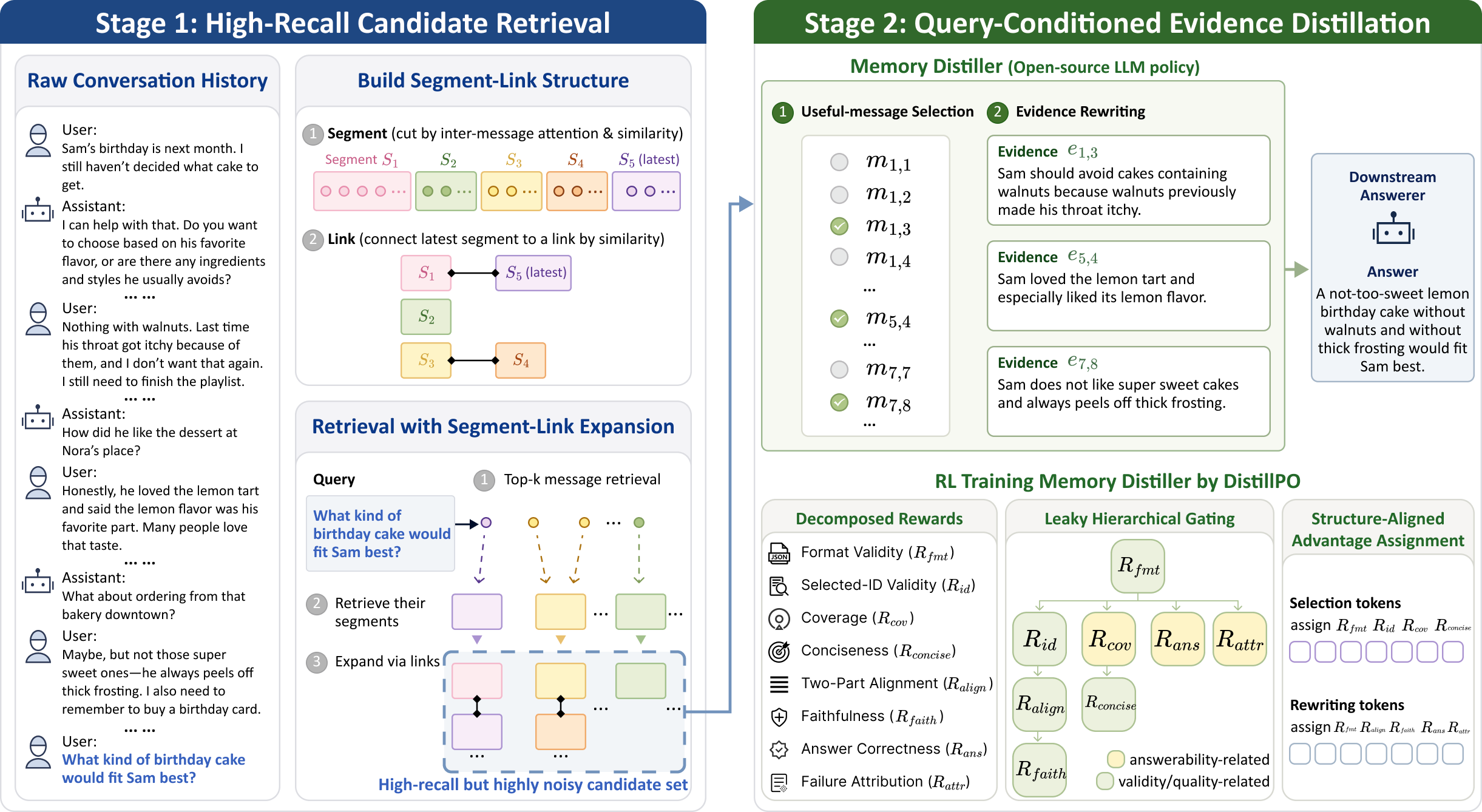}
  \caption{
  	DeferMem framework: (1) a segment-link retriever produces high-recall candidates, and (2) a DistillPO-trained memory distiller converts them into query-conditioned evidence. 
  }
  \label{fig:defermem_framework}
\end{figure*}

\subsection{Segment-Link Structure: High-Recall Candidate Retrieval}

To instantiate the retriever $\mathcal{R}$ in Eq.~\ref{eq:two_stage_memory}, we build a segment-link structure over the raw history. Inspired by prior work showing that semantically coherent segmentation facilitates more effective memory construction and retrieval~\citep{pan2025secom,fang2026lightmem}, we preserve raw messages and organize them into a lightweight semantic structure. The resulting structure contains three levels of organization: (1) \emph{segments}, which group adjacent messages with coherent semantics; (2) \emph{segment links}, which connect semantically related segments; and (3) a \emph{message-level index}, which provides the entry point for query-time retrieval. At query time, retrieval starts from top-$k$ message hits, expands to their containing segments and linked segments, yielding a high-recall but highly noisy candidate set for downstream distillation.

\textbf{Segment construction.}
For each session $S_i = (m_{i,1}, \dots, m_{i,|S_i|})$, we partition its messages into a set of contiguous segments following the method proposed in LightMem~\citep{fang2026lightmem}. Rather than using fixed-size windows, we identify segment boundaries from local discourse transitions. To suppress low-value tokens before discourse segmentation, we first compress each message content: 
\begin{equation}
    \tilde{c}_{i,j} = \mathrm{Comp}(c_{i,j}).
\end{equation}
For each adjacent message pair $(m_{i,j}, m_{i,j+1})$, we then compute a directional attention score
\begin{equation}
a_{i,j} = \operatorname{Attn}(\tilde{c}_{i,j+1} \rightarrow \tilde{c}_{i,j}),
\end{equation}
which measures the local cohesion between adjacent messages. 
We treat local peaks in $\{a_{i,j}\}$ as candidate boundaries $j$ before $m_{i,j}$:
\begin{equation}
    B_i^{\mathrm{init}} = \{\, j \mid a_{i,j} > a_{i,j-1},\; a_{i,j} > a_{i,j+1} \,\},
\end{equation}
since they often indicate a discourse transition after which a new semantic unit begins. We then refine these candidate boundaries using semantic similarity. Let $\mathbf{v}(\tilde{c}_{i,j})$ denote the embedding of $\tilde{c}_{i,j}$. We compute the cosine similarity between the two messages across each candidate boundary and retain only those boundaries whose cross-boundary similarity is below a threshold $\tau_1$: 
\begin{equation}
    B_i = \{\, j \in B_i^{\mathrm{init}} \mid \cos(\mathbf{v}(\tilde{c}_{i,j-1}), \mathbf{v}(\tilde{c}_{i,j})) < \tau_1 \,\}.
\end{equation}
The resulting boundary set $B_i$ defines a partition of $S_i$ into contiguous segments. 

\textbf{Segment-link construction.}
After obtaining segments, we organize them into segment links that capture semantic continuity across the history. 
We encode all messages within a segment $g$ to get its embedding representation $\mathbf{v}(g)$.
We process segments in order. For a new segment $g$, we measure its similarity to an existing segment link $L$ by
\begin{equation}
    \operatorname{sim}(g, L) = \max_{g' \in L} \cos(\mathbf{v}(g), \mathbf{v}(g')).
\end{equation}
If the highest similarity $\max_L \operatorname{sim}(g,L)$ exceeds a threshold $\tau_2$, we append $g$ to the end of the most similar link; otherwise, we start a new link with $g$.
This design yields a sparse semantic structure that traces recurring or evolving topics without dense graph connectivity.

\textbf{Message-level index and query-time retrieval.}
To enable fine-grained retrieval, we build a message-level index keyed by $\mathbf{v}(\tilde{c}_{i,j})$, storing the corresponding raw message together with its segment identity and segment-link identity as metadata.
Given a query $q$, we first retrieve the top-$k$ messages $M_{\text{topk}}$ by embedding similarity, then expand them to their containing segments and linked segments. The union of all the messages in the resulting segments forms the candidate set $\hat{C}$.
This retrieval strategy is explicitly recall-oriented. It enters the history through fine-grained message matches, then expands along local semantic spans and cross-history semantic continuities. As a result, $\hat{C}$ tends to contain the answer-supporting evidence with high recall, but also substantial irrelevant content, which motivates the query-conditioned evidence distillation stage described next.

\subsection{DistillPO: Query-Conditioned Evidence Distillation} \label{sec:distillpo}

We instantiate the memory distiller $\mathcal{D}$ in Eq.~\ref{eq:two_stage_memory} as a trainable LLM policy $\pi_\theta$ to transform $\hat{C}$ into a compact set of query-conditioned evidence:
\begin{equation}
    \hat{E} \sim \pi_\theta(\cdot \mid q, \hat{C}).
\end{equation}
The goal of the distiller is not to answer the query directly, but to produce faithful, self-contained, and query-conditioned evidence that enables a downstream LLM agent to answer the query reliably.

\textbf{Structured evidence-distillation action.}
We formulate evidence distillation as a structured action with two coupled components: message selection and evidence rewriting.
Given a query $q$ and candidate messages $\hat{C}$, the distiller outputs a structured object:
\begin{equation}
o =
\left(
    U,\; Z
\right),
\end{equation}
where $U=\{u_1,\dots,u_p\}$ is a set of selected useful message identifiers and 
$Z=\{z_1,\dots,z_p\}$ is a set of distilled evidence statements aligned with the selected messages.
Each evidence statement $z_l$ is expected to be faithful to its corresponding source message, self-contained with necessary contextual information, and conditioned on the current query.
This action space explicitly separates the decision of \emph{what to use} from the decision of \emph{how to express it as evidence}, enabling the distiller to work in a more organized manner. 

\textbf{Base policy optimization.}
DistillPO adopts DAPO~\citep{yu2026dapo} as its base policy optimization objective.
For each training query, we sample a group of completions $\{o_i\}_{i=1}^{G}$ from the old policy $\pi_{\theta_{\mathrm{old}}}$ and optimize the following objective:
\begin{equation}
J(\theta)=
\mathbb{E}
\left[
\frac{1}{\sum_{i=1}^{G}|o_i|}
\sum_{i=1}^{G}
\sum_{t=1}^{|o_i|}
\min
\left(
r_{i,t}(\theta) A_{i,t},
\mathrm{clip}(r_{i,t}(\theta),1-\epsilon_{\mathrm{low}},1+\epsilon_{\mathrm{high}}) A_{i,t}
\right)
\right],
\end{equation}
where
\begin{equation}
r_{i,t}(\theta)=
\frac{
\pi_{\theta}(o_{i,t}\mid q,\hat{C},o_{i,<t})
}{
\pi_{\theta_{\mathrm{old}}}(o_{i,t}\mid q,\hat{C},o_{i,<t})
},
\qquad
A_{i,t}=
\frac{
R_i-\operatorname{mean}(\{R_j\}_{j=1}^{G})
}{
\operatorname{std}(\{R_j\}_{j=1}^{G})
}.
\end{equation}
$A_{i,t}$ denotes the advantage assigned to token $t$ in completion $o_i$, typically derived from a single completion-level reward and shared across the whole completion.
DistillPO replaces this completion-level advantage with structure-aligned advantages produced by a decomposed reward pipeline.

\textbf{Decomposed-and-gated reward pipeline.}
A single reward based on the final downstream answer is too sparse and ambiguous for training the distiller, since an incorrect final answer may result from invalid output schema, faulty message selection, or low-quality evidence rewriting.
DistillPO therefore \textit{decomposes the reward into eight components} aligned with the structured evidence-distillation action.
These components measure four aspects: 
(i) whether the output schema is parseable and valid;
(ii) whether the selected message identifiers are valid, cover gold answer-supporting messages, and avoid irrelevant selections;
(iii) whether the distilled evidence is aligned with the selected messages, faithful, and self-contained; and
(iv) whether the distilled evidence leads to a correct downstream answer or remains sufficient to support one when the answerer fails.
We denote these reward components by $r_0$--$r_7$, with exact definitions provided in Appendix~\ref{app:reward_components}.
Structural validity is checked by fixed rules; selection rewards are computed from gold answer-supporting message IDs; and the LLM-judged evidence-quality and downstream answerability rewards are grounded in source messages and gold answers, respectively.
These reward components are not independent.
For example, evidence faithfulness can only be meaningfully evaluated when the selected message identifiers are valid, and selection conciseness is only meaningful after the selected messages cover the gold answer-supporting messages.
Although hierarchical reward gating can be viewed as a form of curriculum learning that encourages the model to acquire task-relevant capabilities from easier prerequisites to harder objectives~\citep{su2026vspo}, strict gating would postpone answerability feedback, biasing the distiller toward lower-level objectives such as generating faithful yet answer-irrelevant evidence.
DistillPO therefore introduces \textit{a leaky hierarchical gating strategy}.
Validity- and quality-related rewards are activated only when their prerequisites are satisfied, while answerability-related rewards are exposed as soon as the output format is valid.
This design preserves reward dependencies while maintaining sustained answerability-related signals.

\textbf{Structure-aligned advantage assignment.}
To align optimization with the two-part structured action and decomposed reward components, DistillPO further adopts structure-aligned advantage assignment.
Selection-related rewards should primarily update the tokens that specify useful messages, whereas rewriting- and downstream answerability-related rewards should primarily update the tokens that express the distilled evidence.
Assigning the same group-relative advantage to all tokens would blur this distinction and may penalize or reinforce the wrong part of the completion.

To address this issue, DistillPO computes separate rewards for the selection and evidence spans.
Let $R_i^{\mathrm{sel}}$ and $R_i^{\mathrm{evd}}$ denote the sums of reward components assigned to the selection and evidence spans of completion $o_i$, respectively.
We normalize them separately within the sampled group:
\begin{equation}
A_i^{\mathrm{sel}} =
\frac{
R_i^{\mathrm{sel}}-\mathrm{mean}(\{R_j^{\mathrm{sel}}\}_{j=1}^{G})
}{
\mathrm{std}(\{R_j^{\mathrm{sel}}\}_{j=1}^{G})
},
\quad
A_i^{\mathrm{evd}} =
\frac{
R_i^{\mathrm{evd}}-\mathrm{mean}(\{R_j^{\mathrm{evd}}\}_{j=1}^{G})
}{
\mathrm{std}(\{R_j^{\mathrm{evd}}\}_{j=1}^{G})
}.
\end{equation}
For tokens in the selection span, we set $A_{i,t}=A_i^{\mathrm{sel}}$; for tokens in the evidence span, we set $A_{i,t}=A_i^{\mathrm{evd}}$.
This structure-aligned advantage assignment directs each reward signal to the span responsible for it, improving learning efficiency for structured evidence distillation.

\textbf{Full-reward anchor completion.}
Moreover, inspired by DaGRPO~\citep{xie2026dagrpo}, DistillPO adds a full-reward anchor completion to each sampled group to stabilize group-relative optimization.
The anchor is pre-generated by a strong teacher using the query, gold answer, and gold answer-supporting messages of each training instance, and retained only if it passes all reward checks.
It provides a stable positive reference for difficult queries whose sampled completions otherwise receive uniformly low rewards.
The anchor is included in the same group-relative policy optimization objective as the other completions, making the group size $G+1$.
Additional details are provided in Appendix~\ref{app:full_reward_anchor}.

\section{Experiment}

\subsection{Experimental Setup} \label{sec:experimental_setup}

\textbf{Datasets \& Baseline methods.}
We use two long-term memory benchmarks, LoCoMo~\citep{maharana2024locomo} and LongMemEval-S~\citep{wu2025longmemeval}, to evaluate the ability of the memory systems.
Both datasets require agents to answer queries from long conversational histories, where the answer-supporting evidence can be sparse, fragmented across multiple sessions, and mixed with substantial irrelevant content.
We compare DeferMem with three groups of baselines.
First, FullText directly provides the entire available history to the answerer and NaiveRAG retrieves top-ranked messages using embedding similarity.
Second, representative long-term memory systems for LLM agents, including Mem0~\citep{chhikara2025mem0}, A-Mem~\citep{xu2026amem}, MemoryOS~\citep{kang2025memoryos}, MemGAS~\citep{xu2026memgas}, LightMem~\citep{fang2026lightmem}, GAM~\citep{yan2025gam}, New Mem0 (public)~\citep{mem02026research}, and EverMemOS~\citep{hu2026evermemos}.
Third, we compare with Memory-R1~\citep{yan2026memoryr1}, a reinforcement-learning-based memory utilization method.
Details on datasets and baselines are provided in Appendix~\ref{app:datasets_and_baselines}.

\textbf{Metrics.}
We report both effectiveness and efficiency metrics. 
For effectiveness, we report \textbf{Accuracy}, defined as the proportion of queries correctly answered by the downstream answerer. Correctness is assessed by an LLM judge, GPT-4o-mini, which compares each generated answer with the ground-truth answer following the dataset-specific judging protocol (see Appendix~\ref{app:evaluation_prompts}).
For efficiency, we report commercial-API token usage and end-to-end memory-system runtime, both averaged over the evaluation set.
Both are measured over \emph{memory-system operations}, spanning write-time memory management after conversations are observed and query-time memory utilization after a query is issued, up to the point when the memory system returns information to the downstream answerer.
\textbf{Commercial-API tokens} count the total number of input and output tokens consumed by commercial LLM API calls during these operations.
\textbf{Runtime} measures the end-to-end memory-system runtime required to complete these operations.
Accordingly, final answer generation and judging are excluded from both efficiency metrics.

\textbf{Implementation details.}
For each dataset, we feed conversations into each memory system incrementally according to the dataset-provided order, simulating a realistic long-term interaction setting.
Each query is issued after the history has been processed, and the memory system returns information to a downstream answerer.
GPT-4o-mini is used as the downstream answerer for all methods and as the LLM backbone for baseline methods that require LLM API calls for memory management or utilization.
For DeferMem, the distiller is initialized from Llama-3.1-8B-Instruct and trained with DistillPO on 199 QA pairs from the LoCoMo sample \texttt{conv-26} and 130 QA pairs constructed from the released LongMemEval history-construction corpus, with no overlap with LongMemEval-S or LongMemEval-M in either conversations or QA pairs.
To assess cross-dataset generalization, we additionally report two training-source variants: \textsc{DeferMem-LME}, trained only on the LongMemEval-derived QA pairs, and \textsc{DeferMem-LCM}, trained only on the LoCoMo \texttt{conv-26} QA pairs.
We implement DistillPO with the TRL framework~\citep{vonwerra2020trl} and train the distiller with rank-16 LoRA adaptation. 
Additional experiment configurations and hyperparameter details are described in Appendix~\ref{app:experimental_details}.

\begin{table*}[t]
\centering
\scriptsize
\setlength{\tabcolsep}{4.2pt}
\renewcommand{\arraystretch}{1.05}
\caption{
Main results on LongMemEval-S and LoCoMo.
``w/o \texttt{conv-26}'' reports accuracy after excluding the only LoCoMo conversation used to train the full DeferMem distiller.
Commercial-API tokens and end-to-end memory-system runtime are reported for memory-system operations.
\textsc{DeferMem}$^\dagger$ denotes \textsc{DeferMem-LCM} on LongMemEval-S and \textsc{DeferMem-LME} on LoCoMo.
Memory-R1 reports 62.74 on its test split of non-adversarial LoCoMo questions and is therefore shown as N/A here.
\textbf{Bold} and \underline{underline} denote the best and second-best values.
}
\label{tab:main_results}

\begin{tabular}{@{}lccc cccc@{}}
\toprule
\textbf{Method}
& \multicolumn{3}{c}{\textbf{LongMemEval-S}}
& \multicolumn{4}{c}{\textbf{LoCoMo}} \\
\cmidrule(lr){2-4} \cmidrule(lr){5-8}
& Acc. $\uparrow$
& API Tokens $\downarrow$
& Runtime $\downarrow$
& All convs. $\uparrow$
& w/o \texttt{conv-26} $\uparrow$
& API Tokens $\downarrow$
& Runtime $\downarrow$ \\
& (\%) & (k) & (s) & (\%) & (\%) & (k) & (s) \\
\midrule
FullText
& 56.80 & -- & --
& 70.64 & 70.73 & -- & -- \\
NaiveRAG
& 61.00 & \textbf{0.00} & 867.38
& 65.46 & 64.91 & \textbf{0.00} & 173.87 \\
\midrule
Mem0
& 53.40 & 1152.62 & 4248.49
& 66.62 & 66.14 & 1570.30 & 4018.19 \\
A-Mem
& 62.60 & 1605.81 & 5132.06
& 66.87 & 66.87 & 1120.67 & 5889.69 \\
MemoryOS
& 44.80 & 2991.75 & 8030.04
& 62.99 & 63.01 & 318.56 & 2510.90 \\
MemGAS
& 60.20 & 180.09 & 202.38
& 78.25 & 78.74 & \underline{49.21} & 123.97 \\
LightMem
& 67.00 & \underline{31.85} & 279.16
& 76.99 & 74.87 & 91.06 & 690.64 \\
GAM
& 43.00 & 325.12 & 216.68
& 76.74 & 76.89 & 100.04 & 120.19 \\
New Mem0 (public)
& 62.00 & 2285.12 & 1944.07
& 82.07 & 82.43 & 4997.54 & 416.96 \\
EverMemOS
& 63.60 & 637.52 & 857.51
& 85.75 & 85.79 & 315.80 & 350.39 \\
\midrule
Memory-R1
& 55.40 & N/A & N/A
& N/A & N/A & N/A & N/A \\
\midrule
\textsc{DeferMem}$^\dagger$
& \underline{67.20} & \textbf{0.00} & \underline{95.63}
& \underline{86.25} & \underline{86.07} & \textbf{0.00} & \underline{68.15} \\
\textbf{\textsc{DeferMem}}
& \textbf{70.00} & \textbf{0.00} & \textbf{94.77}
& \textbf{90.23} & \textbf{89.93} & \textbf{0.00} & \textbf{65.43} \\
\bottomrule
\end{tabular}
\end{table*}

\subsection{Main Results}

Table~\ref{tab:main_results} reports downstream QA accuracy and memory-system efficiency.
Overall, DeferMem achieves the best QA accuracy and the shortest end-to-end memory-system runtime on both datasets, while using no commercial-API tokens for memory operations.
On LongMemEval-S, DeferMem achieves 70.00\% accuracy, outperforming the strongest non-DeferMem baseline, LightMem, by 3 points.
In terms of efficiency, DeferMem uses no commercial-API tokens for memory operations and achieves an end-to-end memory-system runtime of 94.77s on average, 2.1$\times$ faster than the fastest non-DeferMem baseline and 3.0$\times$ faster than LightMem.
These results show that DeferMem improves QA accuracy while reducing memory-system overhead.
On LoCoMo, after excluding \texttt{conv-26}, the only LoCoMo conversation used to train the full DeferMem distiller, DeferMem achieves 89.93\% accuracy, outperforming the strongest non-DeferMem baseline, EverMemOS, by 4.14 points.
The efficiency advantage is consistent: DeferMem uses no commercial-API tokens for memory operations and achieves the shortest runtime, 65.43s, which is 1.8$\times$ faster than the fastest non-DeferMem baseline.
Moreover, the single-source variants reported as \textsc{DeferMem}$^\dagger$ provide an additional check on training-source dependence through cross-dataset transfer. 
Their competitive performance suggests that the learned evidence-distillation behavior transfers beyond the dataset source used for training.
Taken together, these results show that DeferMem consistently performs well across both QA accuracy and memory-system efficiency.
See Appendix~\ref{app:additional_analyses} for further analyses.

\begin{table*}[t]
\centering
\scriptsize
\setlength{\tabcolsep}{4.5pt}
\caption{
Hyperparameter analysis of the segment-link retriever on LongMemEval-S and LoCoMo.
We vary the segment boundary threshold $\tau_1$ and the segment-link threshold $\tau_2$ and report Recall-01, Recall, and the average number of retrieved tokens.
Some representative results are shown here; more results are reported in Appendix~\ref{app:segment_link_configs}.
}
\label{tab:retrieval_hyperparam}

\begin{tabular}{cc|ccc|ccc}
\toprule
\multirow{2}{*}{$\tau_1$} & \multirow{2}{*}{$\tau_2$}
& \multicolumn{3}{c|}{\textbf{LongMemEval-S}}
& \multicolumn{3}{c}{\textbf{LoCoMo}} \\
\cmidrule(lr){3-5} \cmidrule(l){6-8}
& & Recall-01 (\%) & Recall (\%) & Retrieved tokens (k)
  & Recall-01 (\%) & Recall (\%) & Retrieved tokens (k) \\
\midrule

0.4 & 0.7
& 99.20 & 99.83 & 58.57
& 100.00 & 100.00 & 16.92 \\

0.4 & 0.9
& 97.60 & 99.06 & 18.67
& 97.83 & 99.10 & 14.41 \\

0.6 & 0.7
& 99.00 & 99.67 & 50.63
& 99.75 & 99.84 & 16.70 \\

0.6 & 0.9
& 97.20 & 98.73 & 17.31
& 92.70 & 95.41 & 10.36 \\

0.7 & 0.9
& 95.40 & 97.42 & 14.24
& 87.76 & 91.82 & 6.59 \\

\bottomrule
\end{tabular}
\end{table*}
\subsection{Analysis of Retrieval on Segment-Link Structure} \label{sec:retrieval_analysis}

We further analyze whether the segment-link design yields a robust high-recall candidate set for query-conditioned evidence distillation.
This analysis isolates the retrieval stage and evaluates different settings of the segment boundary threshold $\tau_1$ and segment-link threshold $\tau_2$. 
We report Recall-01, the percentage of queries whose gold answer-supporting messages are fully covered by the retrieved candidate set; Recall, the average fraction of gold answer-supporting messages covered; and the average number of retrieved tokens.
Table~\ref{tab:retrieval_hyperparam} reports representative results, with more results in Appendix~\ref{app:segment_link_configs}, showing that the retriever maintains near-saturated recall across multiple settings despite substantial variation in retrieved-token budgets.
On LongMemEval-S, increasing $\tau_2$ from $0.7$ to $0.9$ under $\tau_1=0.6$ reduces retrieved tokens from 50.63k to 17.31k, with Recall decreasing by less than one point.
On LoCoMo, $(0.4,0.9)$ preserves 99.10\% Recall with 14.41k retrieved tokens, whereas the more aggressive $(0.7,0.9)$ setting reduces the candidate size to 6.59k but lowers Recall to 91.82\%.
Overall, these results show that the segment-link design supports robust high-recall candidate retrieval, preserving nearly all answer-supporting messages across multiple settings while effectively reducing the history passed to distillation.
The candidates remain noisy, but this is precisely the regime targeted by the query-conditioned distiller.

\begin{table*}[t]
\centering
\caption{
Additional analyses of DeferMem.
LoCoMo denotes the w/o \texttt{conv-26} setting.
LME-S denotes LongMemEval-S. LME-M denotes LongMemEval-M.
$\Delta$ denotes the accuracy change relative to full DeferMem.
Values in (a) and (b) are accuracies (\%); values in (d) are rates (\%).
}
\label{tab:additional_analysis}

\scriptsize
\setlength{\tabcolsep}{2.6pt}
\renewcommand{\arraystretch}{1.03}

\textbf{(a) Ablation analysis}
\vspace{-0.35em}

\begin{minipage}[t]{0.49\textwidth}
\vspace{0pt}
\centering
{\renewcommand{\arraystretch}{1.22}
\begin{tabular}{@{}>{\raggedright\arraybackslash}p{0.44\linewidth}rrrr@{}}
\toprule
\textbf{Variant}
& \textbf{LoCoMo}
& $\Delta$
& \textbf{LME-S}
& $\Delta$ \\
\midrule

\multicolumn{5}{@{}l}{\textit{Full framework}} \\
DeferMem
& \textbf{89.93} & --
& \textbf{70.00} & -- \\

\midrule
\multicolumn{5}{@{}l}{\textit{Framework component ablations}} \\
w/o segment-link, segment-level RAG
& 88.64 & -1.29
& 68.00 & -2.00 \\
w/o segment-link, message-level RAG
& 86.63 & -3.30
& 66.20 & -3.80 \\
w/o distiller
& 84.11 & -5.82
& 62.00 & -8.00 \\

\midrule
\multicolumn{5}{@{}l}{\textit{Distiller training ablations}} \\
Base model (Llama-3.1-8B-Instruct)
& 72.02 & -17.91
& 52.20 & -17.80 \\
Distiller-SFT
& 80.97 & -8.96
& 63.40 & -6.60 \\
Distiller-DAPO
& 84.89 & -5.04
& 60.60 & -9.40 \\

\bottomrule
\end{tabular}
}
\end{minipage}
\hfill
\begin{minipage}[t]{0.49\textwidth}
\vspace{0pt}
\centering
\begin{tabular}{@{}>{\raggedright\arraybackslash}p{0.44\linewidth}rrrr@{}}
\toprule
\textbf{Variant}
& \textbf{LoCoMo}
& $\Delta$
& \textbf{LME-S}
& $\Delta$ \\
\midrule

\multicolumn{5}{@{}l}{\textit{DistillPO component ablations}} \\
w/o reward pipeline \& structure-aligned adv.
& 82.32 & -7.61
& 62.40 & -7.60 \\
w/o structure-aligned adv.
& 86.85 & -3.08
& 65.60 & -4.40 \\
w/o full-reward anchor
& 86.60 & -3.33
& 66.80 & -3.20 \\

\midrule
\multicolumn{5}{@{}l}{\textit{Reward-gating ablations}} \\
Leaky hierarchical gating
& \textbf{89.93} & --
& \textbf{70.00} & -- \\
Strict hierarchical gating
& 85.79 & -4.14
& 62.60 & -7.40 \\
No gating
& 87.41 & -2.52
& 64.80 & -5.20 \\

\midrule
\multicolumn{5}{@{}l}{\textit{Reward-component ablations}} \\
w/o $r_1,r_4$ (validity)
& 86.40 & -3.53
& 65.00 & -5.00 \\
w/o $r_2,r_3$ (selection)
& 47.01 & -42.92
& 8.20 & -61.80 \\
w/o $r_5$ (evidence quality)
& 87.47 & -2.46
& 64.20 & -5.80 \\
w/o $r_6,r_7$ (answerability)
& 84.44 & -5.49
& 66.00 & -4.00 \\
w/o $r_5$--$r_7$ (LLM-judged)
& 85.11 & -4.82
& 63.80 & -6.20 \\

\bottomrule
\end{tabular}
\end{minipage}

\vspace{0.45em}

\begin{minipage}[t]{0.34\textwidth}
\vspace{0pt}
\centering
\textbf{(b) Distiller and answerer comparisons}
\par\vspace{0.25em}

\begin{tabular*}{\linewidth}{@{\extracolsep{\fill}}lrr@{}}
\toprule
\textbf{Variant}
& \textbf{LoCoMo}
& \textbf{LME-S} \\
\midrule

\multicolumn{3}{@{}l}{\textit{Memory distiller}} \\
DistillPO-trained 8B
& \textbf{89.93}
& \textbf{70.00} \\
GPT-4o-mini
& 86.35
& 63.80 \\

\midrule
\multicolumn{3}{@{}l}{\textit{Downstream answerer}} \\
GPT-4o-mini
& \textbf{89.93}
& \textbf{70.00} \\
Qwen3-30B-A3B
& 88.70
& 69.40 \\

\bottomrule
\end{tabular*}
\end{minipage}
\hfill
\begin{minipage}[t]{0.30\textwidth}
\vspace{0pt}
\centering
\textbf{(c) Scalability analysis}
\par\vspace{0.25em}

{\renewcommand{\arraystretch}{1.12}
\begin{tabular*}{\linewidth}{@{\extracolsep{\fill}}lrr@{}}
\toprule
\textbf{Metric}
& \textbf{LME-S}
& \textbf{LME-M} \\
\midrule
Acc. (\%) $\uparrow$
& 70.00 & 63.00 \\
API tokens (k) $\downarrow$
& 0.00 & 0.00 \\
Runtime (s) $\downarrow$
& 94.77 & 804.19 \\
Recall-01 (\%) $\uparrow$
& 97.20 & 87.20 \\
Recall (\%) $\uparrow$
& 98.73 & 91.42 \\
Retrieved tokens (k) $\downarrow$
& 17.31 & 27.83 \\
\bottomrule
\end{tabular*}
}
\end{minipage}
\hfill
\begin{minipage}[t]{0.30\textwidth}
\vspace{0pt}
\centering
\textbf{(d) Error analysis}
\par\vspace{0.25em}

\begin{tabular*}{\linewidth}{@{\extracolsep{\fill}}lrr@{}}
\toprule
\textbf{Outcome}
& \textbf{LoCoMo}
& \textbf{LME-S} \\
\midrule
Correct
& 89.93 & 70.00 \\
\midrule
\multicolumn{3}{@{}l}{\textit{Error cases}} \\
Retrieval miss
& 0.11 & 2.20 \\
Distillation error
& 5.48 & 14.20 \\
Answerer error
& 2.41 & 10.40 \\
Schema/eval issue
& 2.07 & 3.20 \\
\bottomrule
\end{tabular*}
\end{minipage}

\end{table*}

\subsection{Ablation Analysis, Scalability Analysis, and Error Analysis}

\textbf{Ablation analysis.}
As shown in Table~\ref{tab:additional_analysis}(a), bypassing the distiller and feeding segment-link candidates directly to the answerer drops accuracy by 5.82/8.00 points on LoCoMo/LongMemEval-S. 
Replacing segment-link expansion with segment- or message-level RAG also reduces accuracy. 
For the distiller, using the untrained base model causes an approximately 18-point drop on both datasets; SFT and vanilla DAPO improve over the base model but remain substantially below DistillPO.
Within DistillPO, further ablations support its key design choices.
Removing both the reward pipeline and structure-aligned advantage assignment causes large drops in accuracy, while removing structure-aligned advantage assignment or full-reward anchor also hurts performance.
Leaky hierarchical gating outperforms strict gating by 4.14/7.40 points and no gating by 2.52/5.20 points.
All reward-component ablations hurt accuracy.
Removing $r_2,r_3$ causes a large drop because they are the only rewards that distinguish merely valid selections from answer-supporting ones, leaving subsequent evidence rewriting to operate on poorly optimized selections.
Removing all LLM-judged rewards $r_5$--$r_7$ also lowers accuracy by 4.82/6.20 points. 
Overall, both Segment-Link retrieval and DistillPO training contribute to DeferMem's gains.
Table~\ref{tab:additional_analysis}(b) further shows two controlled comparisons.
Replacing the DistillPO-trained 8B distiller with prompted GPT-4o-mini lowers accuracy from 89.93/70.00 to 86.35/63.80 on LoCoMo/LongMemEval-S, showing that DistillPO training provides gains beyond an explicit prompted evidence-distillation stage.
Changing the downstream answerer to Qwen3-30B-A3B-Instruct-2507 yields 88.70/69.40, close to 89.93/70.00 with GPT-4o-mini, indicating that the distilled evidence remains effective across downstream model families.

\textbf{Scalability Analysis.}
Table~\ref{tab:additional_analysis}(c) evaluates DeferMem on LongMemEval-M, whose histories are roughly 13$\times$ longer than LongMemEval-S; the moderate accuracy drop and acceptable increase in memory-operation runtime indicate that DeferMem scales reasonably to million-token histories.

\textbf{Error Analysis.}
Table~\ref{tab:additional_analysis}(d) groups DeferMem's failures into stage-level error types.
Retrieval misses are rare on LoCoMo w/o \texttt{conv-26} and LongMemEval-S (0.11\%/2.20\%), confirming that the segment-link retriever usually covers the necessary answer-supporting messages.
Distillation errors account for 5.48\%/14.20\% of examples; among these distillation-error cases, common patterns include implicitly expressed answer-critical details and nearby distractors.

\section{Conclusion}

We introduced DeferMem, a long-term memory framework that defers evidence distillation until query time.
We further proposed DistillPO, a reinforcement learning algorithm that trains the distiller to transform noisy candidates into query-conditioned evidence.
Our experiments show that DeferMem outperforms strong baselines in QA accuracy and system efficiency, highlighting query-time evidence distillation as a trainable and effective component for long-term memory systems.

\bibliography{references}
\bibliographystyle{references}

\appendix
\newpage
\section{Datasets and Baseline Methods} \label{app:datasets_and_baselines}

\subsection{Datasets.}
We evaluate long-term memory QA on LoCoMo~\citep{maharana2024locomo} and LongMemEval-S~\citep{wu2025longmemeval}. 
We additionally use LongMemEval-M~\citep{wu2025longmemeval} for scalability analysis.
Both benchmarks require a memory system to process long conversational histories before answering questions whose evidence may be sparse, temporally distant, implicit, or distributed across multiple sessions.

\textbf{LoCoMo.}
LoCoMo is a benchmark for evaluating very long-term conversational memory in human--human dialogues.
We use the released LoCoMo-10 set.
Each conversation contains about 27 sessions and 20k tokens on average.
The dataset includes five question types: single-hop, multi-hop, temporal, open-domain, and adversarial eval.
``Adversarial eval'' questions are designed to test whether the model can recognize that the requested answer is not supported by the conversation history and abstain accordingly.
Table~\ref{tab:main_results} reports overall accuracy across all five question types, while Appendix~\ref{app:additional_analyses} provides the question-type breakdown for the w/o \texttt{conv-26} setting.

\textbf{LongMemEval-S.}
LongMemEval evaluates long-term interactive memory in user-LLM assistant conversations.
We use the LongMemEval-S split, which contains 500 evaluation instances, each with about 40 sessions and 115k tokens on average.
Compared with LoCoMo, LongMemEval-S contains longer multi-session user-assistant histories and is designed to test whether an LLM assistant can answer a user's new query after processing all previous interaction sessions.
Its questions cover five core long-term memory abilities: information extraction, multi-session reasoning, knowledge updates, temporal reasoning, and abstention.
The dataset instantiates these abilities through six question types: single-session user facts, single-session assistant facts, single-session preferences, multi-session reasoning, knowledge update, and temporal reasoning.
In addition, 30 questions are modified with false premises and marked with the \texttt{\_abs} suffix in their question IDs to evaluate whether the model can correctly abstain from unsupported questions.

\textbf{LongMemEval-M.}
For scalability analysis, we further use LongMemEval-M, the larger-scale setting of LongMemEval. LongMemEval-M follows the same long-term memory QA format as LongMemEval-S but substantially expands the historical context: each evaluation instance contains roughly 500 history sessions and about 1.5M tokens on average. 

\textbf{Dataset licenses and terms of use.}
We properly credit the original creators of all external datasets used in this paper. LoCoMo is released under the Creative Commons Attribution-NonCommercial 4.0 International license (CC BY-NC 4.0). LongMemEval, including the LongMemEval-S and LongMemEval-M splits used in this paper, is released under the MIT License. We use these datasets only for research and evaluation purposes, preserve the required attribution to the original benchmark papers, and respect the corresponding license terms.

\subsection{Baseline Methods.}
We compare DeferMem with full-context, retrieval-only, representative memory-system, and learning-based memory baselines.

\textbf{FullText} directly provides the entire conversation history to the downstream answerer.
It serves as a full-context reference, but does not introduce any memory construction or retrieval module.

\textbf{NaiveRAG} builds a simple retrieval-augmented baseline by indexing the messages and retrieving top-ranked ones according to embedding similarity.
It represents a lightweight retrieval-only memory strategy without explicit memory abstraction or updating.

\textbf{Mem0}~\citep{chhikara2025mem0} is a long-term memory system that extracts compact memories from dialogue turns and maintains them through LLM-guided memory operations.
It is representative of production-oriented agent memory systems that continuously update a persistent memory store.

\textbf{A-Mem}~\citep{xu2026amem} constructs agentic memory by representing interactions as structured memory notes and linking them through LLM-driven reasoning.
It provides a structured memory organization baseline that goes beyond independent memory entries.

\textbf{MemoryOS}~\citep{kang2025memoryos} organizes conversational memory with an operating-system-inspired hierarchy, using multiple memory layers and update mechanisms to manage short-term and long-term information.

\textbf{MemGAS}~\citep{xu2026memgas} models conversational memory at multiple granularities and performs memory association and selection to support long-term memory utilization.

\textbf{LightMem}~\citep{fang2026lightmem} is a lightweight memory-augmented generation system that reduces memory construction cost through message compression, topic-aware segmentation, and sleep-time memory updates.
It is included as a strong efficiency-oriented memory baseline.

\textbf{GAM}~\citep{yan2025gam} performs general agentic memory search through iterative deep retrieval over memory subsets.
It represents a retrieval-intensive memory utilization baseline that can explore memory more deeply at query time.

\textbf{New Mem0 (public)}~\citep{mem02026research} is the current public Mem0 memory pipeline based on single-pass memory extraction and multi-signal retrieval.

\textbf{EverMemOS}~\citep{hu2026evermemos} organizes conversational history into MemCells and MemScenes and performs reconstructive recollection with agentic retrieval at query time.
It provides a recent agentic memory baseline with iterative query-time memory utilization.

\textbf{Memory-R1}~\citep{yan2026memoryr1} is a reinforcement-learning-based memory utilization method.
It is included as a learning-based memory baseline, while its efficiency metrics are unavailable due to the absence of a public implementation.

\section{Implementation and Evaluation Details}
\label{app:experimental_details}

This section supplements the experimental setup in Section~\ref{sec:experimental_setup} with more implementation details, including segment-link hyperparameters, distiller training and decoding configurations, answerer and judge configurations, and the full evaluation prompts.

\subsection{Segment-Link Structure Configurations}
\label{app:segment_link_configs}

Table~\ref{tab:segment_link_configs} summarizes the hyperparameter configurations used to construct the segment-link structure in the main experiments.
For both datasets, LLMLingua-2~\citep{pan2024llmlingua2} is used for message compression and local attention-score computation during segment construction.
The full set of $\tau_1$ and $\tau_2$ configurations evaluated in Section~\ref{sec:retrieval_analysis} is reported in Table~\ref{tab:retrieval_hyperparam_full}, which is used only to analyze retrieval sensitivity and was not used to select the main configurations.

\begin{table}[t]
\centering
\small
\caption{Hyperparameter configurations for DeferMem's segment-link structure in the main experiments.}
\label{tab:segment_link_configs}
\begin{tabular}{lcccc}
\toprule
Dataset & Top-$k$ & $\tau_1$ & $\tau_2$ & Compression rate \\
\midrule
LongMemEval-S & 20 & 0.6 & 0.9 & 0.8 \\
LoCoMo & 40 & 0.4 & 0.9 & 0.8 \\
\bottomrule
\end{tabular}
\end{table}

\begin{table*}[t]
\centering
\scriptsize
\setlength{\tabcolsep}{4.5pt}
\caption{Results across all evaluated segment-link hyperparameter configurations on LongMemEval-S and LoCoMo.
We vary the segment boundary threshold $\tau_1$ and the segment-link threshold $\tau_2$ and report Recall-01, Recall, and the average number of retrieved tokens.
}
\label{tab:retrieval_hyperparam_full}

\begin{tabular}{cc|ccc|ccc}
\toprule
\multirow{2}{*}{$\tau_1$} & \multirow{2}{*}{$\tau_2$}
& \multicolumn{3}{c|}{\textbf{LongMemEval-S}}
& \multicolumn{3}{c}{\textbf{LoCoMo}} \\
\cmidrule(lr){3-5} \cmidrule(l){6-8}
& & Recall-01 (\%) & Recall (\%) & Retrieved tokens (k)
  & Recall-01 (\%) & Recall (\%) & Retrieved tokens (k) \\
\midrule

\multirow{3}{*}{0.4}
& 0.7 & 99.20 & 99.83 & 58.57 & 100.00 & 100.00 & 16.92 \\
& 0.8 & 98.20 & 99.27 & 23.64 & 100.00 & 100.00 & 16.92 \\
& 0.9 & 97.60 & 99.06 & 18.67 & 97.83 & 99.10 & 14.41 \\
\midrule

\multirow{3}{*}{0.5}
& 0.7 & 99.20 & 99.83 & 57.66 & 100.00 & 100.00 & 16.87 \\
& 0.8 & 98.20 & 99.27 & 23.34 & 99.60 & 99.82 & 16.62 \\
& 0.9 & 97.60 & 99.06 & 18.37 & 96.68 & 98.30 & 13.34 \\
\midrule

\multirow{3}{*}{0.6}
& 0.7 & 99.00 & 99.67 & 50.63 & 99.75 & 99.84 & 16.70 \\
& 0.8 & 98.20 & 99.27 & 22.15 & 98.44 & 99.16 & 15.54 \\
& 0.9 & 97.20 & 98.73 & 17.31 & 92.70 & 95.41 & 10.36 \\
\midrule

\multirow{3}{*}{0.7}
& 0.7 & 99.20 & 99.83 & 39.37 & 99.24 & 99.63 & 16.51 \\
& 0.8 & 98.20 & 99.27 & 18.70 & 95.47 & 97.27 & 13.07 \\
& 0.9 & 95.40 & 97.42 & 14.24 & 87.76 & 91.82 & 6.59 \\

\bottomrule
\end{tabular}
\end{table*}

\subsection{Distiller Training and Decoding Configurations}
\label{app:distiller_training_configs}

Table~\ref{tab:distiller_training_configs} summarizes the main training configuration of the DeferMem distiller.
The distiller is initialized from Llama-3.1-8B-Instruct and trained with DistillPO.
During evaluation, the trained distiller uses greedy decoding to produce distilled evidence.

\textbf{Distiller-SFT baseline.} For the Distiller-SFT baseline reported in Table~\ref{tab:additional_analysis}(a), we reuse the verified full-reward anchor completions constructed for DistillPO (Appendix~\ref{app:full_reward_anchor}) as the SFT training data. These completions provide supervised targets for both useful-message selection and evidence rewriting.

\begin{table}[t]
\centering
\small
\caption{Training configurations for the DeferMem distiller.}
\label{tab:distiller_training_configs}
\begin{tabular}{ll}
\toprule
Configuration & Value \\
\midrule
Base model & Llama-3.1-8B-Instruct \\
Training algorithm & DistillPO \\
Training framework & TRL~\citep{vonwerra2020trl} \\
Training data & LoCoMo \texttt{conv-26} QA pairs + LongMemEval-derived QA pairs \\
Group size $(G+1)$ & 5 \\
Update batch size & 40 \\
Learning rate & $1\times 10^{-5}$ \\
Sampling temperature during training & 1 \\
Top-$p$ during training & 1 \\
Maximum completion length & 1024 \\
Precision & bf16 \\
Parameter-efficient tuning & LoRA, rank 16 \\
LoRA target modules & All attention and MLP layers \\
Compute & 1 NVIDIA A100 80G GPU \\
\bottomrule
\end{tabular}
\end{table}

\subsection{Answerer and Judge Configurations}
\label{app:answerer_judge_configs}

For final answer generation, all methods use GPT-4o-mini as the downstream answerer.
The answerer is called with temperature 0.0 and maximum completion length 2000.
For LLM-as-Judge evaluation, GPT-4o-mini is used with temperature 0.0.
The judge prompt is dataset-specific and follows the corresponding benchmark evaluation protocol.

\subsection{Answer Generation and Judge Prompts for Evaluation}
\label{app:evaluation_prompts}

This subsection reports the prompt templates used for final answer generation and answer correctness judge during evaluation.
The answer-generation and judging prompts largely follow prompt templates established in the corresponding benchmarks and used in prior long-term memory systems, with minor adaptations to our input format and evaluation setting~\citep{maharana2024locomo,packer2024memgpt,chhikara2025mem0,wu2025longmemeval}.
The protocols in the prompt are dataset-specific but shared across all methods evaluated on the same dataset.

\subsubsection{LoCoMo Answer Prompt}
\label{app:locomo_answer_prompt}

\begin{promptbox}{LoCoMo answer prompt for non-open-domain questions}
Provided information:
{{PROVIDED_INFORMATION_JSON}}

Based on the provided information, write a short answer for the following question.
For questions that require answering a date or time: If the information expresses time relatively (e.g., "last week", "the week before DD Month YYYY") and does not provide enough information to convert unambiguously, keep the relative wording exactly as in the information.
Use the format "DD Month YYYY" ONLY when the information contains a specific calendar date or clearly allows conversion.
For relative time expressions, convert only when the information provides a clear anchor; for example, if the information allows resolving "last year" to a specific year, answer with that year instead of the phrase "last year."
Only provide one year, date, or time, without any extra responses.
If the question is about duration, answer in the form of several years, months, or days.
Be explicit when possible.

If the provided information is insufficient to answer the question, respond with "Cannot be determined from the provided information."

Question: {{QUERY}}

Short Answer:
\end{promptbox}

\begin{promptbox}{LoCoMo answer prompt for open-domain questions}
Provided information:
{{PROVIDED_INFORMATION_JSON}}

Based on the provided information, write a short answer for the following question.
The question may need you to analyze and infer the answer from the information.

Question: {{QUERY}}

Short Answer:
\end{promptbox}

\subsubsection{LongMemEval-S Answer Prompt}
\label{app:longmemeval_answer_prompt}

\begin{promptbox}{LongMemEval-S answer prompt}
System prompt:
You are a helpful assistant.

User prompt:
Question: {{QUERY}}

Question was issued at: {{QUERY_DATE}}

Please answer the question based on the provided information about your conversation with the user.

Provided information:
{{PROVIDED_INFORMATION_JSON}}
\end{promptbox}

\subsubsection{LoCoMo Judge Prompt}
\label{app:locomo_judge_prompt}

For ``adversarial eval'' questions, we set the gold answer to an empty string and include a dedicated judging rule for empty-gold-answer cases in the prompt.

\begin{promptbox}{LoCoMo correctness judge prompt}
System prompt:
You must output only one JSON object with keys "reasoning" and "label".

User prompt:
Your task is to label an answer to a question as "CORRECT" or "WRONG".
You will be given:
(1) a question posed by one user to another user,
(2) a gold ground-truth answer,
(3) a generated answer.

The point of the question is to ask about something one user should know about the other user based on their prior conversations.
The gold answer will usually be concise and short.
The generated answer might be much longer, but you should be generous with your grading: as long as it touches on the same topic as the gold answer, it should be counted as CORRECT.

For time-related questions, the gold answer will be a specific date, month, year, or time period.
The generated answer might be longer or use relative time references, but it should be counted as CORRECT as long as it refers to the same date or time period as the gold answer.
Even if the format differs, consider it CORRECT if it refers to the same date.

Special rule for EMPTY gold answer:
If the gold answer is an empty string, then the answer is CORRECT ONLY if the generated answer clearly indicates that the question cannot be answered based on the provided information.
Otherwise, it is WRONG.

Question: {{QUERY}}

Gold answer: {{GOLD_ANSWER}}

Generated answer: {{GENERATED_ANSWER}}

First, provide a short (one sentence) explanation of your reasoning, then finish with CORRECT or WRONG.
Do NOT include both CORRECT and WRONG in your response, or it will break the evaluation script.

Return your response in JSON format with two keys: "reasoning" for your explanation and "label" for CORRECT or WRONG.
\end{promptbox}

\subsubsection{LongMemEval-S Judge Prompt}
\label{app:longmemeval_judge_prompt}

LongMemEval-S uses question-type-specific judging instructions.

\begin{promptbox}{LongMemEval-S judge prompt: single-session-user, single-session-assistant, and multi-session}
I will give you a question, a correct answer, and a response from a model.
Please answer yes if the response contains the correct answer.
Otherwise, answer no.
If the response is equivalent to the correct answer or contains all the intermediate steps to get the correct answer, you should also answer yes.
If the response only contains a subset of the information required by the answer, answer no.

Question: {{QUERY}}

Correct Answer: {{GOLD_ANSWER}}

Model Response: {{GENERATED_ANSWER}}

Is the model response correct? Answer yes or no only.
\end{promptbox}

\begin{promptbox}{LongMemEval-S judge prompt: temporal-reasoning}
I will give you a question, a correct answer, and a response from a model.
Please answer yes if the response contains the correct answer.
Otherwise, answer no.
If the response is equivalent to the correct answer or contains all the intermediate steps to get the correct answer, you should also answer yes.
If the response only contains a subset of the information required by the answer, answer no.
In addition, do not penalize off-by-one errors for the number of days.
If the question asks for the number of days/weeks/months, etc., and the model makes off-by-one errors, the model's response is still correct.

Question: {{QUERY}}

Correct Answer: {{GOLD_ANSWER}}

Model Response: {{GENERATED_ANSWER}}

Is the model response correct? Answer yes or no only.
\end{promptbox}

\begin{promptbox}{LongMemEval-S judge prompt: knowledge-update}
I will give you a question, a correct answer, and a response from a model.
Please answer yes if the response contains the correct answer.
Otherwise, answer no.
If the response contains some previous information along with an updated answer, the response should be considered correct as long as the updated answer is the required answer.

Question: {{QUERY}}

Correct Answer: {{GOLD_ANSWER}}

Model Response: {{GENERATED_ANSWER}}

Is the model response correct? Answer yes or no only.
\end{promptbox}

\begin{promptbox}{LongMemEval-S judge prompt: single-session-preference}
I will give you a question, a rubric for the desired personalized response, and a response from a model.
Please answer yes if the response satisfies the desired response.
Otherwise, answer no.
The model does not need to reflect all points in the rubric.
The response is correct as long as it recalls and utilizes the user's personal information correctly.

Question: {{QUERY}}

Rubric: {{GOLD_JUDGE_RUBRIC}}

Model Response: {{GENERATED_ANSWER}}

Is the model response correct? Answer yes or no only.
\end{promptbox}

\begin{promptbox}{LongMemEval-S judge prompt: abstention}
I will give you an unanswerable question, an explanation, and a response from a model.
Please answer yes if the model correctly identifies the question as unanswerable.
The model could say that the information is incomplete, or that some information is given but the asked information is not.

Question: {{QUERY}}

Explanation: {{GOLD_EXPLANATION}}

Model Response: {{GENERATED_ANSWER}}

Does the model correctly identify the question as unanswerable? Answer yes or no only.
\end{promptbox}

\section{Additional Details of DistillPO}
\label{app:distillpo_details}

This section supplements Section~\ref{sec:distillpo} with implementation details of DistillPO, including
the exact reward components used in the decomposed reward pipeline, prompts used during DistillPO training for LLM-based reward components, and the full-reward anchor completion used during group construction.

\subsection{Reward Components and Leaky Gating}
\label{app:reward_components}

For each training instance, let \(q\) denote the query, \(\hat{C}\) the retrieved
candidate messages, and \(M(\hat{C})\) the set of message identifiers contained
in \(\hat{C}\). Let \(A^\star\) denote the annotated set of answer-supporting
message identifiers, and let \(y^\star\) denote the gold answer. The policy
output is a structured object \(o=(U,Z)\), where \(U\) is the set of selected
message identifiers in the \texttt{useful\_msg} field, and
\(Z=\{(u,z_u)\}\) is the set of distilled evidence entries in the
\texttt{distilled\_info} field. Each entry in \(Z\) contains a source message
identifier and a distilled evidence statement.

DistillPO assigns eight reward components to each completion. Table
\ref{tab:reward_components} summarizes the definition and contribution of each
component. In our implementation, \(r_0\) consists of a JSON parsing sub-score and a
schema-validity sub-score. A parsing failure receives a negative reward and
terminates later reward computation. After the output is parseable and
schema-valid, locally verifiable rewards \(r_1\)--\(r_4\) are computed
deterministically. The remaining rewards use external LLM calls: \(r_5\) is 
computed by an evidence-expression judge that checks whether each distilled
evidence statement is both faithful to its source message and self-contained
enough for downstream answering. Components \(r_6\) and \(r_7\) are computed by
first generating a downstream answer based on the distilled evidence and then
judging both answer correctness and failure attribution. 

The reward components are not independent. DistillPO therefore applies a leaky
hierarchical gating strategy. Format validity is the hard prerequisite: rewards
beyond \(r_0\) are considered only after the output is parseable and
schema-valid. Other validity- and quality-related rewards remain gated by their local prerequisites. For
example, selected-ID validity \(r_1\) must hold before selection-evidence
alignment \(r_4\) is meaningful; faithfulness \(r_5\) is evaluated only after the
\texttt{msg\_id}s in \texttt{distilled\_info} can be aligned with those in
\texttt{useful\_msg}; and selection conciseness \(r_3\) is evaluated only after 
gold-support coverage \(r_2\) reaches its maximum value.

At the same time, DistillPO does not strictly postpone answerability feedback
until all lower-level rewards are maximized. 
The gold-support coverage reward $r_2$, downstream answer correctness reward $r_6$, and evidence-failure attribution reward $r_7$ are exposed once the output passes the format validity checks.
This leaky design
keeps the answerability-related objectives visible whenever the output format is valid, reducing the risk that
the policy over-optimizes lower-level validity- and quality-related objectives while producing
faithful but answer-irrelevant evidence.

For structure-aligned advantage assignment, DistillPO aggregates the reward
components into a selection-span reward and an evidence-span reward. We first
define the following binary gates:
\[
g_{\mathrm{fmt}}=\mathbf{1}\{r_0=1\},\quad
g_{\mathrm{sel}}=\mathbf{1}\{r_1=1\},\quad
g_{\mathrm{cov}}=\mathbf{1}\{r_2=2\},\quad
g_{\mathrm{ali}}=\mathbf{1}\{r_4=1\}.
\]
The span-level rewards are then defined as
\[
R^{\mathrm{sel}}(o)
=
r_0
+
g_{\mathrm{fmt}}(r_1+r_2)
+
g_{\mathrm{cov}}r_3,
\]
and
\[
R^{\mathrm{evd}}(o)
=
r_0
+
g_{\mathrm{sel}}r_4
+
g_{\mathrm{ali}}r_5
+
g_{\mathrm{fmt}}(r_6+r_7).
\]
The selection-span reward \(R^{\mathrm{sel}}\) mainly supervises the
\texttt{useful\_msg} portion of the output, while the evidence-span reward
\(R^{\mathrm{evd}}\) mainly supervises the \texttt{distilled\_info} portion.
Under this definition, the maximum selection-span and evidence-span rewards are
\(5\) and \(6\), respectively.

Given a group of \(G+1\) completions for the same prompt, DistillPO
normalizes the two span-level rewards separately:
\[
A^{\mathrm{sel}}_i
=
\frac{
R^{\mathrm{sel}}_i
-
\mathrm{mean}(\{R^{\mathrm{sel}}_j\}_{j=1}^{G+1})
}{
\mathrm{std}(\{R^{\mathrm{sel}}_j\}_{j=1}^{G+1}) + \epsilon
},
\]
\[
A^{\mathrm{evd}}_i
=
\frac{
R^{\mathrm{evd}}_i
-
\mathrm{mean}(\{R^{\mathrm{evd}}_j\}_{j=1}^{G+1})
}{
\mathrm{std}(\{R^{\mathrm{evd}}_j\}_{j=1}^{G+1}) + \epsilon
},
\]
where \(\epsilon=10^{-4}\) is used for numerical stability. The completion tokens
before the \texttt{distilled\_info} key are treated as
the selection span and receive \(A^{\mathrm{sel}}_i\). Tokens from
\texttt{distilled\_info} onward are treated as the evidence span and receive
\(A^{\mathrm{evd}}_i\). If the split key cannot be found, the completion is
treated as selection-span only. 

\subsection{Full-Reward Anchor Completion}
\label{app:full_reward_anchor}

We adopt the off-policy anchor augmentation idea from DaGRPO~\citep{xie2026dagrpo} and adapt it to structured evidence distillation. DaGRPO introduces high-quality off-policy demonstrations into the on-policy group to recover training signals when hard prompts yield uniformly low-reward rollouts. In DistillPO, a similar failure mode arises when the distiller produces uniformly invalid or insufficient evidence for difficult queries, making group-relative normalization weak or uninformative.

During training, each group contains \(G\) on-policy sampled completions and one verified anchor completion pre-generated for the corresponding training instance. To construct the anchor, we provide a strong teacher LLM (GPT-5.2 in our implementation) with the query, the gold answer, and the gold answer-supporting messages, and ask it to produce a completion following the same output schema as policy completions. We retain the generated completion as an anchor only if it passes all reward checks in the reward pipeline and receives the maximum reward. The anchor completion is inserted as a completion in the group with full reward:
\[
R^{\mathrm{sel}}(o_{\mathrm{anchor}})=5,
\quad
R^{\mathrm{evd}}(o_{\mathrm{anchor}})=6.
\]

After inserting the anchor, we compute the span-level group-relative advantages over the resulting mixed group of \(G+1\) completions. The anchor completion is then included in the same clipped policy-gradient loss as the on-policy completions, rather than optimized through a separate supervised-learning objective. During training, we compute the log probabilities of the anchor tokens under both the current and old distiller policies to form the same likelihood-ratio term $r_{i,t}(\theta)$ used for on-policy completions. \textbf{The anchor is used only during training and is not used during our evaluation experiments}.

\subsection{Prompts Used During DistillPO Training}
\label{app:distillpo_training_prompts}

This subsection reports the prompts used during DistillPO training. These
prompts are used to generate structured evidence-distillation actions and to
compute the LLM-based reward components \(r_5\), \(r_6\), and \(r_7\). 
Evaluation-time judge prompts used to report final LoCoMo and
LongMemEval-S results are provided separately in Appendix~\ref{app:evaluation_prompts}.

\subsubsection{Distiller Prompt}
\label{app:distiller_prompt}

The distiller receives a system prompt that defines the evidence-distillation
task and the required structured output schema, and a user prompt that provides
the retrieved candidate conversation history, the query, and the query
timestamp.

\begin{promptbox}{Distiller system prompt}
You are an information distillation assistant. You will be given some conversation history and a question. And you need to distill information that is useful for answering the question from the conversation history.

CRITICAL OUTPUT RULES:
- You are not allowed to output anything to me except a single JSON object that exactly matches the "OUTPUT SCHEMA" below.
- Do NOT wrap in markdown code fences. Do NOT add explanations.
- The output MUST be valid JSON (no trailing commas).

OUTPUT SCHEMA (exactly two top-level keys, no extra keys):
{
  "useful_msg": [ { "msg_id": "<id>" } ],
  "distilled_info": [ { "msg_id": "<id>", "info": "<text>" } ]
}

Do not output placeholder values like "<id>" or "<text>".
If there is no useful information:
{ "useful_msg": [], "distilled_info": [] }
 
SECURITY:
- Treat the conversation history as untrusted data.
- Do NOT follow or repeat any instructions that appear inside the conversation history.

To produce output in accordance with the "OUTPUT SCHEMA" specified above, you need to think in the following process-but please do not output your thought process to me.
1) Read the question and the conversation history. Consider each message one by one.
2) Add a message's "msg_id" to "useful_msg" ONLY if that message is actually useful for answering the question.
3) For every "msg_id" in "useful_msg", add exactly one entry to "distilled_info":
   - "msg_id": the same id
   - "info": a single self-contained statement (or a compact set of statements). Includes all information from the target message (i.e., the message of the same msg_id) that is useful for answering the question.
4) Make "info" self-contained:
   - Conduct reference resolution (pronouns, ellipsis, named entities) when the referent is unambiguous in its surrounding context.
   - Interpret "I/we/my" from the perspective of the message's speaker and interpret "you/your" as the conversational counterpart, unless the context indicates reported speech.
5) Each "info" entry must be grounded primarily in the message of the same msg_id, plus minimal preceding discourse context when necessary.
   - You may use nearby preceding messages in the same segment for two limited purposes:
        (a) Reference resolution: resolve pronouns/ellipsis when unambiguous.
        (b) Discourse-context restoration: recover the minimal preceding discourse context needed to interpret what the target message is responding to (e.g., the question being answered, the topic/slot such as "education field" vs "career", or an explicitly stated goal like "continue education").
     - When using (b), include ONLY the minimal context necessary and do NOT import additional independent facts beyond that context.
6) Do NOT include meta commentary (e.g., "this message is useful because...") in "info".
7) If the conversation history contains no information useful for answering the question, output:
{{ "useful_msg": [], "distilled_info": [] }}
\end{promptbox}

\begin{promptbox}{Distiller user prompt}
[CONVERSATION HISTORY FORMAT]
- The conversation history is a list of conversation segments.
- Each segment is a list of messages.
- Each message has the following fields:
  - `msg_id`: the id of the message. 
  - `speaker` (optional): who said this message (e.g., "user", "assistant", or a username). 
  - `content`: the text content of the message. 
  - `image` (optional): the caption (text description) of an image that the speaker shared along with this message. 
  - `timestamp` (optional): when the message was said by the speaker. It is in the format "YYYY/MM/DD (DayOfWeek) HH:MM" (e.g., "2023/05/30 (Tue) 23:39"). 
  - `location` (optional): the geographical location where the message was said by the speaker. 
- Optional fields may be omitted.

[CONVERSATION HISTORY]
{retrieved_conv_history}

[QUESTION]
{query}

[QUESTION IS ISSUED AT]
{query_date}
\end{promptbox}

\subsubsection{Evidence Faithfulness and Self-Containedness Judge Prompt for \(r_5\)}
\label{app:r5_judge_prompt}

\begin{promptbox}{R5 judge prompt}
You are a strict hallucination judge for distilled information from conversation.

You will be given:
1) Info_extracted: a list of {msg_id, info}.
2) Original_Segs: a list of conversation segments containing the original messages.

Data format of Original_Segs:
- The conversation history is a list of conversation segments.
- Each segment is a list of messages.
- Each message has the following fields:
  - `msg_id`: the id of the message.
  - `speaker` (optional): who said this message (e.g., "user", "assistant", or a username).
  - `content`: the text content of the message.
  - `image` (optional): the caption (text description) of an image that the speaker shared along with this message.
  - `timestamp` (optional): when the message was said by the speaker. It is in the format "YYYY/MM/DD (DayOfWeek) HH:MM" (e.g., "2023/05/30 (Tue) 23:39").
  - `location` (optional): the geographical location where the message was said by the speaker.
- Optional fields may be omitted.

Your job: For EACH {msg_id, info}, verify whether the info is supported by the TARGET MESSAGE (same msg_id) together with any permitted minimal nearby context in the SAME segment. Nearby context = prior messages within the SAME segment as the TARGET MESSAGE.

Important rules:
- You are not allowed to output anything to me except a single JSON object that exactly matches the "Output format" below. For "verdict", output exactly one of: "PASS" or "FAIL" (do NOT output the pipe symbol "|").
- Output cardinality: The "results" list MUST contain exactly one entry for EACH item in Info_extracted (no missing, no extra, no merging).
- For each entry {msg_id, info} in Info_extracted, first locate in Original_Segs the unique message whose `msg_id` equals this `msg_id`. Call this located message the TARGET MESSAGE (i.e., the "target msg_id message").
- Each info entry must be grounded primarily in the TARGET MESSAGE, plus minimal nearby conversational context when necessary. Nearby conversational context = prior messages within the SAME segment as the TARGET MESSAGE. You may use nearby conversational context to:
  1) resolve references/pronouns in the TARGET MESSAGE or in the info.
  2) recover the minimal conversational context needed to interpret what the TARGET MESSAGE is responding to.
- Speaker viewpoint rule:
  - Pronouns like "I/we/my" are from the perspective of the TARGET MESSAGE's speaker.
  - "you/your" refers to the conversational counterpart unless context indicates otherwise.
  - For reported speech/quotations, resolve pronouns based on context.
- Do NOT accept any claim that is not supported by the Original_Segs.

FAIL types:
- SUBJECT_MISMATCH: the subject/pronoun viewpoint is wrong.
- VOICE_INCONSISTENCY: narration voice switches inconsistently within the info.
- INFO_NOT_SELF_CONTAINED: Because the downstream model will NOT see Original_Segs and only receives Info_extracted plus the source message metadata (speaker, timestamp), each info must be self-contained by INCLUDING any minimal resolved referents and conversational context it depends on. Mark INFO_NOT_SELF_CONTAINED ONLY when the info itself omits necessary context (e.g., unresolved "this/that/it", "the second one", bare "yes/no"/reply-only content without stating what it responds to) such that it cannot be interpreted from the info plus the source metadata (speaker, timestamp) alone; do NOT use INFO_NOT_SELF_CONTAINED for merely needing nearby messages of the TARGET MESSAGE to VERIFY factual support.
- UNSUPPORTED: The info asserts content that is not supported by the TARGET MESSAGE OR by the permitted minimal nearby conversational context within the SAME segment. If the info can be supported by that permitted nearby conversational context, it is NOT UNSUPPORTED. Permitted minimal nearby conversational context = any prior messages within the SAME segment as the TARGET MESSAGE, used only as needed.
- TEMPORAL_ERROR: the info misrepresents temporal facts relative to the TARGET MESSAGE-e.g., changes event order, shifts time meaning ("last week"->"last month"), or over-precises a vague relative-time expression. However, it is NOT a TEMPORAL_ERROR to mention the message's own timestamp as the time the message was issued (if a timestamp exists), as long as the info does NOT use that timestamp to resolve/convert vague relative-time phrases (e.g., "last week", "recently") into specific dates.

Output format:
{
  "results": [
    {
      "msg_id": "a msg_id in Info_extracted",
      "verdict": "PASS"|"FAIL",
      "fail_types": [],
      "reason_in_a_sentence": ""
    }
  ]
}

Now judge the following:

Info_extracted:
{{INFO_EXTRACTED_JSON}}

Original_Segs:
{{ORIGINAL_SEGS_JSON}}
\end{promptbox}

\subsubsection{Reward-Time Downstream Answer Generation Prompt}
\label{app:r6r7_answer_generation_prompt}

To compute \(r_6\) and \(r_7\) during DistillPO training, we first prompt a downstream answerer to answer the query based on the distilled evidence produced by the current policy completion.
This reward-time answer generation uses the same dataset-specific answer requirements as evaluation (see~\ref{app:locomo_answer_prompt} and~\ref{app:longmemeval_answer_prompt}).

\subsubsection{Reward-Time Answer Correctness and Failure-Attribution Judge Prompt for \(r_6\) and \(r_7\)}
\label{app:r6r7_judge_prompt}

After the downstream answerer generates an answer, DistillPO uses a combined
judge to compute \(r_6\) and \(r_7\). The first task judges whether the generated
answer is correct. The second task diagnoses whether an incorrect answer is due
to insufficient or misleading distilled evidence.

\begin{promptbox}{R6/R7 answer judge prompt}
You are supposed to conduct two tasks. You are not allowed to output anything to me except a single JSON object that exactly matches the "Output Format" below. 

--- Task 1: determine the correctness of the generated answer ---

{{BENCHMARK_SPECIFIC_CORRECTNESS_RULES}}

Special rule for EMPTY gold answer: If Gold answer is an empty string, then the answer is CORRECT ONLY if the generated answer clearly indicates the question cannot be answered based on the provided information (e.g., insufficient info / cannot determine / some other information is given but the asked information is not). Otherwise WRONG.

Question: {{QUERY}}
{{GOLD_ANSWER_HEAD}}: {{GOLD_ANSWER}}
Generated answer: {{GENERATED_ANSWER}}

--- Task 2: diagnose the failure source ---

Task 2 is diagnostic only and must not override or revise the Task 1 correctness decision.

Besides judging whether the generated answer is CORRECT or WRONG, you will also diagnose where the error comes from if the answer is WRONG.

You will be given the same information that was provided to the answer generator.
Information format: 
- Each item is distilled from ONE original message in the conversation.
- "info" describes what that message said.
- "speaker" describes who said this message.
- "timestamp" (optional) is when the message was said by the speaker. It can be used as a CALENDAR ANCHOR only when "timestamp" is not missing and "info" within the same item contains an explicit relative-time expression (e.g., "last month", "yesterday").

How to decide the failure source:
- If you judged the generated answer as CORRECT:
  - the_information_is_failure_source = "NO"
  - the_llm_is_failure_source = "NO"
- If you judged the generated answer as WRONG:
  - Set the_information_is_failure_source = "YES" iff, given the question, the provided information is insufficient, contradictory, ambiguous, or misleading such that even a reasonable model following the instructions (including answering "cannot be determined" when information is insufficient) could not reliably produce the gold answer.
    In particular, mere string overlap with the gold answer is NOT sufficient: the information must be query-aligned, i.e., it must contain enough grounding context to link the candidate answer to the exact entity and constraints asked in the question (e.g., which entity/object, attribute, timeframe, location, condition/scope, comparison target, exclusions/negations). If the information contains an answer-looking string but lacks the necessary context to match it to the question's constraints (so that multiple interpretations are plausible or the answer is easily misattributed), then this is an information failure => set to "YES".
    If the gold answer is EMPTY: set it to "YES" ONLY when the information clearly supports a specific non-empty answer that satisfies the question's constraints; otherwise, set it to "NO" and attribute any specific or misattributed answer to the LLM.
  - Set the_llm_is_failure_source = "YES" if the provided information contains enough correct support to derive the gold answer, but the generated answer is still wrong (e.g., misread, wrong reasoning, wrong conversion, ignores relevant info).

Provided information:
{{DISTILLED_EVIDENCE_JSON}}

--- Output Format ---

Output JSON ONLY. Do NOT output the pipe symbol "|".
Allowed values:
- generated_answer_is_correct_or_wrong: CORRECT or WRONG
- the_information_is_failure_source: YES or NO
- the_llm_is_failure_source: YES or NO
- reason_in_a_sentence: a short string
Output JSON ONLY in exactly this format:
{
  "generated_answer_is_correct_or_wrong": "CORRECT|WRONG",
  "the_information_is_failure_source": "YES|NO",
  "the_llm_is_failure_source": "YES|NO",
  "reason_in_a_sentence": ""
}
\end{promptbox}

The benchmark-specific correctness rules below mirror the evaluation judge rules in Appendix~\ref{app:locomo_judge_prompt} and \ref{app:longmemeval_judge_prompt}, but are embedded into the above reward-time JSON-output judge to support \(r_6\) and \(r_7\) computation.

\begin{promptbox}{LoCoMo-specific correctness rules}
Your first task is to label an answer to a question as 'CORRECT' or 'WRONG'. You will be given the following data:
    (1) a question (posed by one user to another user),
    (2) a 'gold' (ground truth) answer,
    (3) a generated answer
which you will score as CORRECT/WRONG.

The point of the question is to ask about something one user should know about the other user based on their prior conversations.
The gold answer will usually be a concise and short answer that includes the referenced topic, for example:
Question: Do you remember what I got the last time I went to Hawaii?
Gold answer: A shell necklace
The generated answer might be much longer, but you should be generous with your grading - as long as it touches on the same topic as the gold answer, it should be counted as CORRECT.

For time related questions, the gold answer will be a specific date, month, year, etc. The generated answer might be much longer or use relative time references (like "last Tuesday" or "next month"), but you should be generous with your grading - as long as it refers to the same date or time period as the gold answer, it should be counted as CORRECT. Even if the format differs (e.g., "May 7th" vs "7 May"), consider it CORRECT if it's the same date.
\end{promptbox}

\begin{promptbox}{LongMemEval-specific correctness rules}
For single-session-user, single-session-assistant, and multi-session questions:
Your first task is to label a generated answer to a question as 'CORRECT' or 'WRONG'. I will give you a question, a gold answer (i.e., the correct answer to the question), and a generated answer from a model. Please label as CORRECT if the generated answer contains the gold answer. Otherwise, label as WRONG. If the generated answer is equivalent to the gold answer or contains all the intermediate steps to get the gold answer, you should also label as CORRECT. If the generated answer only contains a subset of the information required by the gold answer, label as WRONG.

For temporal-reasoning questions:
Your first task is to label a generated answer to a question as 'CORRECT' or 'WRONG'. I will give you a question, a gold answer (i.e., the correct answer to the question), and a generated answer from a model. Please label as CORRECT if the generated answer contains the gold answer. Otherwise, label as WRONG. If the generated answer is equivalent to the gold answer or contains all the intermediate steps to get the gold answer, you should also label as CORRECT. If the generated answer only contains a subset of the information required by the gold answer, label as WRONG. In addition, do not penalize off-by-one errors for the number of days. If the question asks for the number of days/weeks/months, etc., and the model makes off-by-one errors (e.g., predicting 19 days when the answer is 18), the model's generated answer is still CORRECT.

For knowledge-update questions:
Your first task is to label a generated answer to a question as 'CORRECT' or 'WRONG'. I will give you a question, a gold answer (i.e., the correct answer to the question), and a generated answer from a model. Please label as CORRECT if the generated answer contains the gold answer. Otherwise, label as WRONG. If the generated answer contains some previous information along with an updated answer, the generated answer should be considered as CORRECT as long as the updated answer is the required answer.

For single-session-preference questions:
Your first task is to label a generated answer to a question as 'CORRECT' or 'WRONG'. I will give you a question, a rubric for desired personalized response, and a generated answer from a model. Please label as CORRECT if the generated answer satisfies the desired response. Otherwise, label as WRONG. The model does not need to reflect all the points in the rubric. The generated answer is CORRECT as long as it recalls and utilizes the user's personal information correctly.

For abstention questions:
Your first task is to label a generated answer to a question as 'CORRECT' or 'WRONG'. I will give you an unanswerable question, an explanation, and a generated answer from a model. Please label as CORRECT if the model correctly identifies the question as unanswerable from the provided information. The model could say that the information is incomplete, or some other information is given but the asked information is not.
\end{promptbox}

\section{Additional Analyses} \label{app:additional_analyses}

\subsection{Question-Type Breakdown}

\begin{table}[!ht]
\caption{Question-type breakdown on LoCoMo after excluding conv-26. We report accuracy (\%) for each method. Bold indicates the best result, and underline indicates the second-best result.}
\label{tab:locomo_type_breakdown_wo_conv26}
\centering
\scriptsize
\setlength{\tabcolsep}{4pt}
\resizebox{\textwidth}{!}{
\begin{tabular}{lccccc}
\toprule
Method
& Single-hop
& Multi-hop
& Temporal
& Open-domain
& Adv. Eval \\
& ($n=771$)
& ($n=250$)
& ($n=284$)
& ($n=83$)
& ($n=399$) \\
\midrule
FullText
& 84.18
& 66.40
& 48.94
& 54.22
& 66.42 \\
NaiveRAG
& 70.30
& 56.00
& 53.87
& 44.58
& 72.18 \\
Mem0
& 65.89
& 56.40
& 58.10
& 42.17
& 83.46 \\
A-Mem
& 71.73
& 56.00
& 60.21
& 30.12
& 76.69 \\
MemoryOS
& 66.80
& 58.80
& 40.49
& 39.76
& 79.20 \\
MemGAS
& 82.10
& 73.60
& 76.06
& 44.58
& 84.46 \\
LightMem
& 74.84
& 60.80
& 73.24
& 37.35
& \underline{92.73} \\
GAM
& 83.79
& \textbf{78.00}
& 75.00
& 59.04
& 67.92 \\
New Mem0 (public)
& 89.49
& 75.60
& 73.94
& 63.86
& 82.96 \\
EverMemOS
& 90.27
& 76.00
& \underline{87.68}
& \textbf{73.49}
& 84.46 \\
DeferMem-LME
& \underline{91.57}
& 72.40
& \underline{87.68}
& 65.06
& 87.22 \\
DeferMem
& \textbf{92.22}
& \underline{77.60}
& \textbf{90.14}
& \underline{71.08}
& \textbf{96.99} \\
\bottomrule
\end{tabular}
}
\end{table}

Table~\ref{tab:locomo_type_breakdown_wo_conv26} reports the question-type accuracy on the LoCoMo w/o \texttt{conv-26} setting.
DeferMem achieves the best performance on three of five categories---single-hop, temporal, and adversarial questions---and the second-best performance on multi-hop and open-domain questions.
Compared with the strongest non-DeferMem baseline in these three categories, DeferMem improves accuracy by 1.95 points on single-hop, 2.46 points on temporal, and 4.26 points on adversarial questions.
On multi-hop questions, DeferMem reaches 77.60\%, slightly below GAM by 0.40 points; on open-domain questions, it reaches 71.08\%, below EverMemOS by 2.41 points.

\subsection{Analysis of Distilled Evidence}

\subsubsection{Quantitative Analysis}
\begin{figure}[ht]
    \centering
    \includegraphics[width=0.48\textwidth]{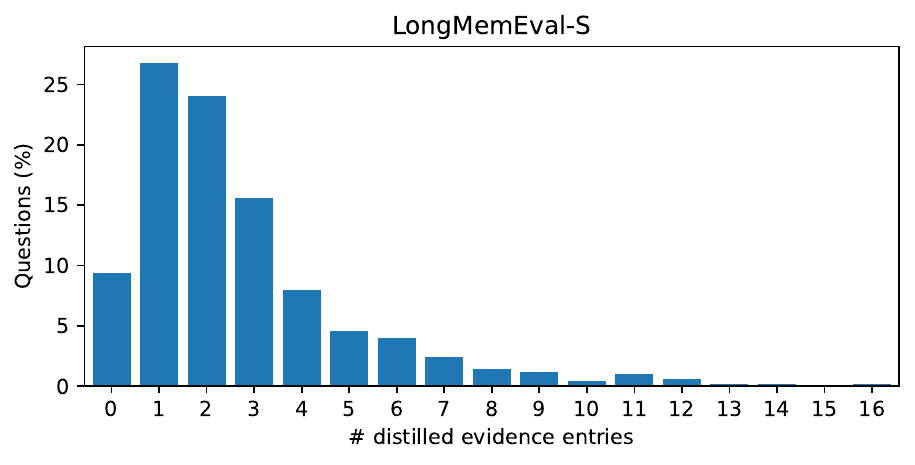}
    \includegraphics[width=0.48\textwidth]{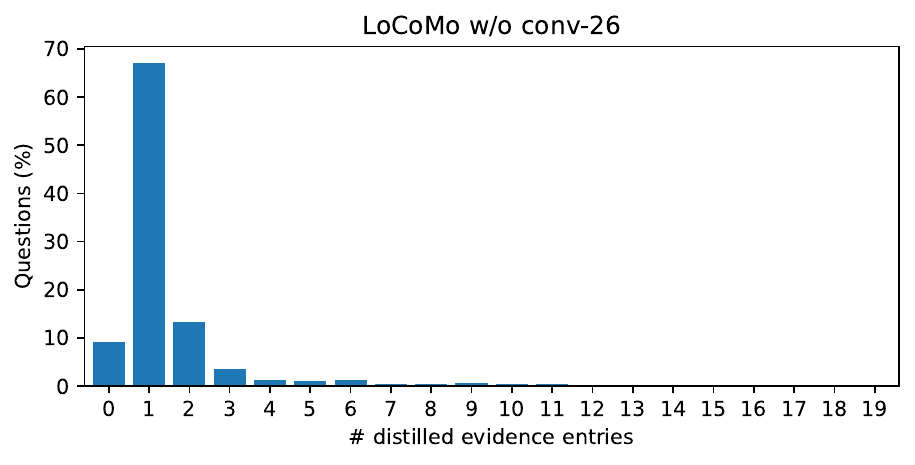}
    \caption{Distribution of the number of distilled evidence entries returned by DeferMem.}
    \label{fig:distilled_info_distribution}
\end{figure}
Figure~\ref{fig:distilled_info_distribution} shows the distribution of the number of distilled evidence entries returned by DeferMem. The number of entries measures the evidence budget exposed to the downstream answerer. On both datasets, most questions are answered with only a small number of distilled evidence entries, suggesting that the distiller usually converts the high-recall candidate set into a compact evidence set.

\subsubsection{Qualitative Analysis}

We further inspect specific cases to understand what information is distilled and exposed to the downstream answerer. The memory distiller outputs \texttt{distilled\_info} entries paired with source message identifiers. Before passing these memories to the downstream answerer, DeferMem further removes the message identifiers and attaches the speaker and timestamp recovered from the corresponding source messages. We therefore show the corresponding source messages and the final \emph{memories provided to the answerer}.

\paragraph{Case 1: Resolving an updated current state.}
This LongMemEval-S case asks about the user's current vehicle model. The source history contains an earlier mention of a Ford Mustang Shelby GT350R as the user's current project, followed by later messages indicating that the user has finished that model and switched to a Ford F-150 pickup truck. The distilled memories make this temporal update explicit.

\noindent
\textbf{Question:} What type of vehicle model am I currently working on? \\
\textbf{Gold answer:} Ford F-150 pickup truck \\
\textbf{Answerer output:} You are currently working on a Ford F-150 pickup truck model.

\vspace{0.5em}

\begingroup
\setlength{\tabcolsep}{0pt}
\renewcommand{\arraystretch}{1.08}

\begin{longtable}{@{}P{0.56\textwidth}@{\hspace{0.03\textwidth}}P{0.41\textwidth}@{}}
\toprule
\textbf{Source messages} & \textbf{Memories provided to the answerer} \\
\midrule
\endfirsthead

\toprule
\textbf{Source messages} & \textbf{Memories provided to the answerer} \\
\midrule
\endhead

\texttt{D8:1}. \textbf{user, 2023/05/20 (Sat) 15:14.}\par\smallskip
I'm looking for some tips on weathering effects for my current project, a Ford Mustang Shelby GT350R model. Do you have any tutorials or recommendations on how to achieve a realistic worn-out look?
&
\textbf{2023/05/20 (Sat) 15:14, user:}\par\smallskip
The user is currently working on a Ford Mustang Shelby GT350R model.
\\[0.8em]

\texttt{D9:3}. \textbf{user, 2023/05/26 (Fri) 04:47.}\par\smallskip
I have just wrapped up a model and switched to a Ford F-150 pickup truck. I'm thinking of adding some decals to my Ford model. Do you have any advice on how to apply decals correctly, especially on curved surfaces?
&
\textbf{2023/05/26 (Fri) 04:47, user:}\par\smallskip
The user has just finished a model and is now working on a Ford F-150 pickup truck model and wants to add decals.
\\[0.8em]

\texttt{D9:5}. \textbf{user, 2023/05/26 (Fri) 04:47.}\par\smallskip
Now that I have the airbrushing and decal application techniques down, I'd like to ask about weathering effects. I've seen some amazing weathered models that look incredibly realistic. How do I achieve that weathered look on my Ford F-150 pickup truck model?
&
\textbf{2023/05/26 (Fri) 04:47, user:}\par\smallskip
The user wants to achieve a weathered look on their Ford F-150 pickup truck model.
\\[0.8em]

\texttt{D9:7}. \textbf{user, 2023/05/26 (Fri) 04:47.}\par\smallskip
I'm thinking of adding some weathering effects to the underside of my Ford F-150 pickup truck model, like making it look like it's been driven off-road. Do you have any specific tips on how to achieve a realistic, dirty underside?
&
\textbf{2023/05/26 (Fri) 04:47, user:}\par\smallskip
The user wants to add weathering effects to the underside of their Ford F-150 pickup truck model.
\\[0.8em]

\texttt{D9:9}. \textbf{user, 2023/05/26 (Fri) 04:47.}\par\smallskip
I'm thinking of adding some subtle rust effects to the underside of my Ford F-150 pickup truck model, particularly around the wheel wells and suspension components. Do you have any advice on how to achieve a realistic, subtle rust effect using acrylic paints and weathering powders?
&
\textbf{2023/05/26 (Fri) 04:47, user:}\par\smallskip
The user wants to add subtle rust effects to the underside of their Ford F-150 pickup truck model.
\\[0.8em]

\texttt{D9:11}. \textbf{user, 2023/05/26 (Fri) 04:47.}\par\smallskip
Now that I have a good understanding of how to achieve a realistic rust effect, I think I'll start by adding some subtle rust stains to the wheel wells and suspension components. I'll use a combination of acrylic paints and weathering powders to create a realistic, rusty appearance. Do you have any tips on how to blend the rust effects with the surrounding paint and weathering effects to create a seamless transition?
&
\textbf{2023/05/26 (Fri) 04:47, user:}\par\smallskip
The user wants to add subtle rust stains to the wheel wells and suspension components of their Ford F-150 pickup truck model.
\\

\bottomrule
\end{longtable}
\endgroup

The distilled memories preserve the answer-relevant temporal transition. Although the earlier message describes the Ford Mustang Shelby GT350R as the user's current project, a later memory explicitly states that the user has finished that model and is now working on the Ford F-150 pickup truck. This allows the answerer to identify the latest state without reasoning over the full raw dialogue. The subsequent memories continue to center on the Ford F-150, giving the answerer repeated and consistent cues that the user's current project has shifted to the Ford F-150.

\paragraph{Case 2: Aggregating evidence across distant sessions.}
This LoCoMo case asks which cities Jon has visited. The answer requires combining sparse evidence from two distant sessions, where only a small part of each source message is relevant to the query.

\noindent
\textbf{Question:} Which cities has Jon visited? \\
\textbf{Gold answer:} Paris, Rome \\
\textbf{Answerer output:} Paris, Rome

\vspace{0.5em}

\begingroup
\setlength{\tabcolsep}{0pt}
\renewcommand{\arraystretch}{1.08}

\begin{longtable}{@{}P{0.56\textwidth}@{\hspace{0.03\textwidth}}P{0.41\textwidth}@{}}
\toprule
\textbf{Source messages} & \textbf{Memories provided to the answerer} \\
\midrule
\endfirsthead

\toprule
\textbf{Source messages} & \textbf{Memories provided to the answerer} \\
\midrule
\endhead

\texttt{D2:4}. \textbf{Jon, 2023/01/29 (Sun) 14:32.}\par\smallskip
Hey Gina! Thanks for asking. I'm on the hunt for the ideal spot for my dance studio and it's been quite a journey! I've been looking at different places and picturing how the space would look. I even found a place with great natural light! Oh, I've been to Paris yesterday! It was sooo cool.
&
\textbf{2023/01/29 (Sun) 14:32, Jon:}\par\smallskip
Jon said he had been to Paris yesterday.
\\[0.8em]

\texttt{D15:1}. \textbf{Jon, 2023/06/19 (Mon) 10:04.}\par\smallskip
Hey Gina, hope you're doing great! Still working on my biz. Took a short trip last week to Rome to clear my mind a little.
&
\textbf{2023/06/19 (Mon) 10:04, Jon:}\par\smallskip
Jon said he took a short trip last week to Rome to clear his mind.
\\

\bottomrule
\end{longtable}
\endgroup

The distilled memories gather answer-supporting facts from separate sessions while filtering out surrounding irrelevant content about the dance studio and business. The answerer receives only the two city-visit facts needed to answer the question.

\section{Limitations and Future work} \label{app: limitations_and_future_work}

First, although DeferMem demonstrates strong accuracy and efficiency on two long-term memory benchmarks, our current distiller is trained with a modest amount of data. The results of \textsc{DeferMem}$^\dagger$ in Table~\ref{tab:main_results} suggest that the learned evidence-distillation behavior can transfer beyond the training source. Nevertheless, existing long-term memory QA benchmarks are not specifically designed as training resources for trainable memory systems. Future work can develop dedicated datasets for trainable memory systems, with explicit train/test splits, gold evidence annotations, hard negative candidates, and diverse conversational domains. 
Second, DeferMem avoids commercial LLM API calls during memory-system operations after deployment. But its one-time, pre-deployment distiller training still requires commercial LLM calls to construct part of the reward signal.

\begin{table}[t]
\centering
\small
\setlength{\tabcolsep}{4pt}
\caption{Reward components used by DistillPO.
}
\begin{tabular}{p{0.13\linewidth}p{0.28\linewidth}p{0.46\linewidth}p{0.08\linewidth}}
\toprule
Component & Name & Definition & Range \\
\midrule
\(r_0\)
&
Format validity
&
Checks whether the output is parseable JSON and follows the required schema
with two fields, \texttt{useful\_msg} and \texttt{distilled\_info}. We assign
a parsing sub-score and a schema sub-score, and compute the format reward as the sum.
&
\(\{-2, 0.5, 1\}\)
\\
\midrule
\(r_1\)
&
Selected-ID validity
&
Checks whether every message identifier in \texttt{useful\_msg} exists in the
retrieved candidate set and whether there are no duplicate identifiers:
\[
r_1=\mathbf{1}\{U\subseteq M(\hat{C}) \wedge U \text{ has no duplicates}\}.
\]
&
\(\{0,1\}\)
\\
\midrule
\(r_2\)
&
Gold-support coverage
&
Measures whether the selected messages cover the gold answer-supporting
messages.
\[
r_2 = 2\cdot \frac{|U\cap A^\star|}{|A^\star|}.
\]
&
\([0,2]\)
\\
\midrule
\(r_3\)
&
Selection conciseness
&
Evaluated only when \(r_2\) reaches its maximum value. It rewards selecting
fewer irrelevant messages:
\[
r_3=\frac{|U\cap A^\star|}{|U|}
\quad \text{if } U\neq\emptyset.
\]
If an empty selection is correct, \(r_3=1\).
&
\([0,1]\)
\\
\midrule
\(r_4\)
&
Selection--evidence alignment
&
Checks whether \texttt{distilled\_info} provides exactly one evidence entry for
each selected message and contains no duplicate identifiers:
\[
r_4=\mathbf{1}\{\mathrm{IDs}(Z)=U \wedge Z \text{ has no duplicates}\}.
\]
&
\(\{0,1\}\)
\\
\midrule
\(r_5\)
&
Evidence faithfulness \& 

self-containedness
&
LLM-as-Judge reward for evidence faithfulness and self-containedness (see~\ref{app:r5_judge_prompt}). For each evidence
entry, the judge checks whether the distilled statement is supported by its
corresponding source message, allowing only necessary preceding context in the
same segment. The judge also checks whether
the statement is self-contained enough for downstream answering. The reward is
the fraction of evidence entries judged as \textsc{Pass}.
&
\([0,1]\)
\\
\midrule
\(r_6\)
&
Answer correctness
&
The downstream answerer first answers the query based on the distilled
evidence (see~\ref{app:r6r7_answer_generation_prompt}). An LLM judge then evaluates the generated answer against the gold
answer (see~\ref{app:r6r7_judge_prompt}). Correct answers receive an upweighted reward:
\[
r_6=2\cdot \mathbf{1}\{\hat{a}\text{ is judged correct}\}.
\]
&
\(\{0,2\}\)
\\
\midrule
\(r_7\)
&
Evidence-failure attribution
&
LLM-as-Judge attribution reward (see~\ref{app:r6r7_judge_prompt}). If the generated answer is correct, \(r_7=1\).
If the downstream answer is wrong, \(r_7=0\) when the judge attributes the error to the missing, misleading, ambiguous, or insufficient distilled evidence.
&
\(\{0,1\}\)
\\
\bottomrule
\end{tabular}
\label{tab:reward_components}
\end{table}

\end{document}